\documentclass[conference]{IEEEtran}
\usepackage{times}

\usepackage[numbers]{natbib}
\usepackage[T1]{fontenc}
\usepackage{multicol}
\usepackage[bookmarks=true]{hyperref}
\usepackage{amsmath,amssymb,bm}
\usepackage{setspace}
\usepackage[Symbol]{upgreek}
\usepackage{graphicx,subfigure}
\usepackage{graphics}
\usepackage{subfloat}
\usepackage{floatrow} 
\usepackage{multirow}
\usepackage{array}
\usepackage{enumerate}
\usepackage{sidecap}
\usepackage[all]{xy}
\usepackage{authblk}
\usepackage{comment}
\usepackage{changepage}   
\usepackage{wrapfig}
\usepackage{url}
\usepackage{tabularx,pbox,booktabs,caption}
\usepackage{flushend}
\usepackage{anyfontsize}
\usepackage{float}
\usepackage{todonotes}
\usepackage[percent]{overpic}
\graphicspath{{./pics/}}
\usepackage{amssymb}
\usepackage{dblfloatfix}    

\usepackage{amsfonts}
\usepackage[T1]{fontenc}

\usepackage{mathabx}
\usepackage{upgreek}
\usepackage{algpseudocode}
\usepackage{algorithm}

\usepackage{amsthm}
\newtheorem{definition}{\textbf{Definition}}

\algnewcommand\algorithmicforeach{\textbf{for each}}
\DeclareMathOperator*{\argmax}{arg\,max}
\DeclareMathOperator*{\argmin}{arg\,min}
\algdef{S}[FOR]{ForEach}[1]{\algorithmicforeach\ #1\ \algorithmicdo}
\algnewcommand\algorithmicinput{\textbf{Input:}}
\algnewcommand\INPUT{\item[\algorithmicinput]}
\algnewcommand\algorithmicoutput{\textbf{Output:}}
\algnewcommand\OUTPUT{\item[\algorithmicoutput]}

\title{
Reachable Space Characterization of Markov Decision Processes with Time Variability
}

\IEEEoverridecommandlockouts
\author{Junhong Xu$^{\dagger}$, Kai Yin$^{\dagger}$, Lantao Liu
\thanks{$^{\dagger}$ Authors contributed equally. }
\thanks{J. Xu and L. Liu are with 
the School of Informatics, Computing, and Engineering  at Indiana University, Bloomington, IN 47408, USA. E-mail:
{\tt\small \{xu14, lantao\}@iu.edu}.
K. Yin is with HomeAway, Inc. E-mail:
{\tt\small yinkai1000@gmail.com}. }
}

\begin{document}

\maketitle
\thispagestyle{empty}
\pagestyle{empty}

\begin{abstract}
We propose a solution to a time-varying variant of Markov Decision Processes which can be used to address decision-theoretic planning problems for autonomous systems operating in unstructured outdoor environments.
We explore the time variability property of the planning stochasticity and investigate the state reachability, based on which we then develop an efficient iterative method that offers a good trade-off between solution optimality and time complexity.  
The reachability space is constructed by analyzing the means and variances of states' reaching time in the future. 
We validate our algorithm through extensive simulations using ocean data, and the results show that our method achieves a great performance in terms of both solution quality and computing time.

\end{abstract}

\section{INTRODUCTION}


Autonomous vehicles that operate in the air and water are 
easily disturbed by stochastic environmental forces such as 
turbulence and currents.
The motion planning in such uncertain environments can be 
modeled using a \textit{decision-theoretic planning} framework where the substrate is the Markov Decision Processes (MDPs)~\cite{sutton2018reinforcement} and its partially observable variants~\cite{jaakkola1995reinforcement}.  

To cope with various sources of uncertainty, the prevalent methodology for addressing 
the decision-making of autonomous systems typically takes advantage of 
the known characterization such as probability distributions for vehicle motion uncertainties.
However, 
the characterization of the uncertainty can vary in time
with some extrinsic factors related to environments 
such as the time-varying disturbances in oceans as shown in Fig.~\ref{fig:dynamic_ocean}.
Despite many successful achievements~\cite{van2012motion, luo2016importance, he2010puma, martinez2007active, huang2018learning}, existing work typically does not synthetically integrate environmental variability with respect to time into the decision-theoretic planning model. 



Although it can be easily recognized that the time-varying stochastic properties represent a more general description of the uncertainty \cite{QMSM2018},  addressing the related planning problems is not an incremental extension to the basic time-invariant counterpart methods.  This is because many components in the basic time-invariant model become time-varying terms simultaneously, requiring substantial re-modeling work and a completely new solution design. 
Therefore, in this work we explore the time variability of the planning stochasticity and investigate the state reachability, 
and these important properties allow us to gain insights for devising an efficient approach with a good trade-off between the solution optimality and the time complexity.
The reachability space is  constructed essentially by analyzing the first and second moments of expected future states' reaching time. 
Finally, we validate our method in the scenario of navigating a marine vehicle in the ocean, and our simulation results show that the proposed approach produces results close to the optimal solution but requires much smaller computational time comparing to other baseline methods.

{
\begin{figure}
    \centering
    \includegraphics[width=0.9\textwidth]{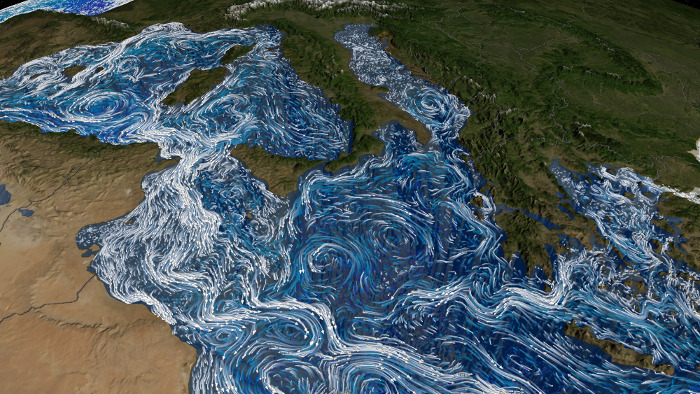}
    \vspace{-0.5pt}
    \caption{\small A motivating scenario: the ocean environment is highly dynamic. Autonomous vehicle planning needs to consider the strong time-varying disturbances caused by ocean currents. (Source: NASA Scientific Visualization Studio.) \vspace{-10pt}}
    \label{fig:dynamic_ocean} 
\end{figure}
}


\section{RELATED WORK}





Most existing methods that solve MDPs in the literature assume 
time-invariant transition models~\cite{sutton2018reinforcement}. 
Such an assumption has also been used in most literature on reinforcement learning (RL)~\cite{Sutton1998,DayanRL08,Bagnell_2013_7451}, 
where an agent tries to maximize accumulated rewards by interacting with its environment. 
A technique related to temporal analysis is temporal difference (TD) learning~\cite{Sutton88learningto,Tesauro1995TDL}. 
RL extends the TD technique by constantly correcting previous predictions based on the availability of new observations~\cite{Sutton1998,Boyan99least-squarestemporal}. 
Unfortunately, existing TD techniques are typically built on time-invariant models. 
Research on time variability has also been conducted in multi-agent co-learning scenarios where multiple agents learn in the environment simultaneously and enable 
the transition of the environment to evolve over time~\cite{he2016opponent,zheng2018deep,everett2018learning}. 
To tackle the environmental uncertainty, multiple stationary MDPs have been utilized to obtain better environmental representations~\cite{banerjee2017quickest,hadoux2014sequential,da2006dealing}. 
Policies are learned for each MDP and then switched if a context change is detected. 
In general, these approaches use a piece-wise stationary approximation, i.e., a time-invariant MDP is applied in each time period of a multi-period horizon.


Instead of directly modeling time-varying transition functions, exogenous variables outside of MPDs may be leveraged to characterize the variation of transition functions ~\cite{li2018faster}.
Some relevant work employs partially observable MDPs to deal with non-stationary environments~\cite{choi2000environment, doshi2015bayesian}, 
or uses Semi-Markov Decision Processes (SMDPs) to incorporate continuous action duration
time in the model~\cite{Sutton99betweenmdps,puterman2014markov, jianyong2004average, beutler1986time}.
However, these frameworks still essentially assume time-invariant transition models. 

Recently, the time-dependent MDP has been analyzed where 
the space-time states are directly employed~\cite{boyan2001exact}.
It has been shown that even under strong assumptions and restrictions, 
the computation is still prohibitive~\cite{RachelsonFG09}.
Relevant but different from this model, 
the time-varying MDP 
is formulated by directly incorporating time-varying transitions whose distribution can be approximated or learned from environments~\cite{liu2018solution}. 
This method is compared with our proposed algorithm in the experiment section. 

Reachability analysis has been vastly researched in the control community where the majority of work falls in non-stochastic domains. 
For example, an important framework utilizes 
Hamilton-Jacobi reachability analysis to guarantee 
control policies to drive the dynamical systems within some pre-defined safe set of states under bounded disturbances~\cite{mitchell2005time, bansal2017hamilton}. 
Control policies can also be learned through machine learning approaches to keep the system outside of unsafe states~\cite{gillula2012guaranteed, gillula2013reducing}. 
In addition, convex optimization 
procedures are carried out to compute reachable funnels within which the states must remain under some control policy~\cite{majumdar2017funnel}.
These funnels are then used to compute robot motions online with a safety guarantee. 



\section{Problem Formulation}

We are motivated by problems that autonomous vehicles (or manipulators) move 
toward defined goal states under exogenous time-varying disturbances. These problems can be modeled as a time-varying Markov Decision Process.
\subsection{Time-Varying Markov Decision Process}\label{sec:tvmdp_intro}

We represent a time-varying Markov Decision Process (TVMDP) as a 5-tuple  $(\mathbb{S}, \mathbb{T}, \mathbb{A}, \mathcal{T}, R)$.  
The \textit{spatial} state space $\mathbb{S}$ and the action space $\mathbb{A}$ are finite.
We also need an extra discrete {\em temporal} space $\mathbb{T}$ in our discussions, though our framework supports continuous time models. 
Because the MDP now is time-varying, 
an action must depend on both state and time; 
that is, for each state $s\in\mathbb{S}$ and each time $t\in\mathbb{T}$, we have a feasible action set $A(s, t)\subset\mathbb{A}$. The entire set of feasible state-time-action triples is $\mathbb{F} := \{(s, t, a)\in \mathbb{S}\times\mathbb{T}\times\mathbb{A}\}$. 
There is a probability transition law $\mathcal{T}(s, t, a; \cdot, \cdot )$ on $\mathbb{S}\times\mathbb{T}$ for all $(s, t, a)\in \mathbb{F}$. 
Thus, $\mathcal{T}(s, t, a; s', t')$ specifies the probability of transitioning to state $s'$ at a future time $t'$ given the current state $s$ and time $t$ ($t\leq t'$) under action $a$.
The final element $R: \mathbb{F}\rightarrow\mathbb{R}^1$ is a real-valued reward function that depends on the state-time-action 
triple.

Comparing with the classic MDP representation, TVMDP contains an additional time space $\mathbb{T}$;  
the transition law $\mathcal{T}$ and the reward function depend on the state, action, and time.
Therefore, the major difference between TVMDPs and classic MDPs is that the transition probability and reward function are time-dependent. 

We consider the class of deterministic Markov policies~\cite{puterman2014markov}, denoted by $\Pi$; that is, the mapping $\pi\in\Pi: \mathbb{S}\times\mathbb{T}\rightarrow\mathbb{A}$ depends on the current state and the current time, and $\pi(s, t)\in A(s, t)$ for a given state-time pair. 
The initial state $s_0\in\mathbb{S}$ at time $t=0$ and policy $\pi\in\Pi$ determine a stochastic trajectory (path) process $\tau:=\{(s_t, a_t)\}_{t\geq 0}$. 
The two terms, path and trajectory, will be used interchangeably. 
For a given initial state $s_0$ and starting time $t_0$, the expected discounted total reward is represented as:
\begin{equation}
    v^\pi(s_0, t_0)=\mathbb{E}^\pi_{s_0, t_0}\left[\sum_{k=0}^{\infty} \gamma^kR(s_k, t_k, a_k)\right],
\end{equation}
where $\gamma\in[0, 1)$ is the discount factor that discounts the reward at a geometric decaying rate. 
Our aim is to find a policy $\pi^*$ to maximize the total reward from the starting time $t_0$ at the initial state $s_0$, i.e.,
\begin{equation}
    \pi^*(s_0, t_0)=\argmax_{\pi\in\Pi} v^\pi(s_0, t_0).
\end{equation}
Accordingly, the optimal value is denoted by $v^*(s_0, t_0)$.

\subsection{Passage Percolation Time}\label{sec:PPT}
The 2D plane in which the autonomous system operates is modeled as the spatial state space. 
The plane is discretized into grids, and the center of each grid represents a state.
Two spatial states are connected if their corresponding grids are neighbors, 
i.e. a vehicle in one grid is able to transit to the other grid directly without passing through any other grids.
Because the vehicle motion follows physical laws (e.g., motion kinematics and dynamics), 
travel time is required for the vehicle to transit between two different states.
Let $h(s, t, s')$ be the local transition time for a vehicle to travel from state $s$ at time $t$ to a connected state $s'$. 
Such a local transition time is time-dependent and is assumed deterministic. 

If, however, the two states $s$ and $s'$ are not connected, 
then the transition time between them is a random variable and depends on the trajectory of the vehicle.
For any finite path $\tau=\{(s, a_0), ..., (s', a_n)\}_{n > 0}$ starting from time $u$ at state $s$ and ending with state $s'$,
we define the \textit{Passage Percolation Time} (PPT) between $s$ and $s'$ to be 
\begin{equation}
H_{s, s'}^{u} = \sum_{(s_{k}, a_{k})\in\tau} h(s_k, t_k, s_{k+1}), \label{PPT}
\end{equation}
where $t_0=u, t_k=t_{k-1} + h(s_{k-1}, t_{k-1}, s_k)$ and $0< k < n$. In addition, we require $H_{s, s}^{u}=0$.
By definition it is the transition time (travel time) on a path from the state $s$ at time $u$ until firstly reaching the state $s'$ \cite{auffinger201550, grimmett1999percolation}. 
If the local transition time $h$ does not depend on time, the definition of Eq.\eqref{PPT} is exactly the same as the conventional {\em passage time} for percolation \cite{auffinger201550}. 
We would like to emphasize that $H^u_{s, s'}$ is a random variable, which relies on the realized path between $s$ and $s'$. Under the policy $\pi$, the mean and variance of the PPT are assumed to exist, and are denoted by $\mu^{\pi}_{s,u;s'}:=\mathbb{E}^\pi[H^u_{s, s'}]$ and  $(\sigma^{\pi}_{s,u;s'})^2=\mathbb{V}^\pi[H^u_{s, s'}]$, respectively. 

\subsection{Spatiotemporal State Space Representation}\label{sec: ST-Space}
One can 
view a TVMDP as a classic MDP by 
defining the product of both 
\textit{spatial} and {\em temporal} spaces  $\mathbb{S}\times\mathbb{T}$ 
as a new state space $\mathbb{S}'$. 
Namely, the state space now stands for the spatiotemporal space in our context. 
In this representation, one 
can imagine that the spatial state space $\mathbb{S}$ is duplicated 
on every discrete time ``layer" to form a collection of spatial states along the temporal dimension as
$\{\mathbb{S}_1, \mathbb{S}_2, ...\}$, where each $\mathbb{S}_k$ is the same as $\mathbb{S}$. The state-time pair $(s, t)$ corresponds to a state in this spatiotemporal space. 
Transition links are added by concatenating states on different time layers, constrained by the local transition time $h$. 
Similar spatiotemporal representation is also adopted in many other fields \cite{shang2019integrating, boyan2001exact, MAHMOUDI201619}.

We emphasize here that the discrete time intervals and transition links between states in the spatiotemporal representation can be determined by the underlying motion kinematics of autonomous vehicles via the local transition time. This is a time discretization-free mechanism and is naturally supported by our proposed TVMDP framework. We will show an example in Section~\ref{sec:realistic}.


\subsection{TVMDP Value Iteration in Spatiotemporal Space}\label{sec: spatio_temporal}


The optimal policy $\pi^*$ for TVMDPs may be conceptually achieved by the conventional value iteration approach through sweeping the entire state space $\mathbb{S}$ and time space $\mathbb{T}$. 
The TVMDP value iteration (VI) amounts to 
iterating
the following state-time value function until convergence,
\begin{equation}
    v(s, t) = \max_{a\in A(s, t)} \left\{ R(s, t, a) + \gamma\cdot\mathbb{E}[v(s', t')\mid s, a; t]\right\}, \label{ST-Valuefunction}
\end{equation}
where $s'$ is the next state to visit at time $t'$ from the current state $s$ at time $t$, and 
\begin{equation*}
    \mathbb{E}[v(s', t')\mid s, a; t] = \sum_{s'\in\mathbb{S}, t'\in\mathbb{T}}\mathcal{T}(s, t, a; s', t')\cdot v(s', t')
\end{equation*}
is the weighted average of the value functions of all the next possible spatiotemporal states.

The value function Eq.~\eqref{ST-Valuefunction} is a modification of the conventional Bellman equation as it includes a notion of time. 
In addition to propagating values spatially from next states, it also backs up the value temporally from a future time. 
Moreover, the benefits of the spatiotemporal representation in applications are readily seen, 
as the solution to Eq.~\eqref{ST-Valuefunction} is equivalent to applying dynamic programming directly to the spatiotemporal state space. 

A typical spatiotemporal state space is very large, 
especially when high state resolution is needed. 
The naive dynamic programming-based value iteration or policy iteration involves backing up state-values not only from the spatial dimension but also from the temporal dimension. 
It is generally intractable to solve for the exact optimal policy due to the so-called curse of dimensionality~\cite{liu2016mdp}. 


\section{Methodology}
In this section, we present tractable iterative algorithms for TVMDP by a reduction of the spatiotemporal state space in each iteration. 
Our approach is grounded in characterizing the most 
possibly reachable set of states through the first and second moments of the passage percolation time. 

\subsection{Overview}\label{sec:method}
One of the major challenges to solving the Bellman equation (Eq.~\eqref{ST-Valuefunction}) is the search in a large spatiotemporal state space. 
Once we are able to reduce the whole spatiotemporal space to a tractable size, it is then possible to obtain solutions by, for instance, the value iteration algorithm, within a reasonable time span. 
Given a policy, an initial state-time pair, and a probability transition law, 
for a fixed (spatial) state $s$, 
probabilities of visiting $s$ at different times are highly likely different.
If we are able to quantify the reachability of spatiotemporal states by visiting probability, and trim the whole spatiotemporal space by removing those with small reachability, 
it is highly likely that the optimal total reward will not be affected much (under certain 
restrictions on the variability of the reward and transition functions), 
and we can gain significant benefits from reducing the computation. 

The previously introduced variable Passage Percolation Time (PPT) $H^u_{s,s'}$ sheds 
light on estimating the chances of reaching state $s'$ by evaluating the transition time 
from a state $s$ at time $u$. 
Although the exact probability distribution of PPT is generally hard to obtain, 
its first and second moments are relatively easy to compute. 
Therefore, {\em we use the expectation and variance of PPT between the initial spatial state $s_0$ and another arbitrary state $s$ to characterize the most possible time interval to reaching $s$.} 
Following this way, 
we are able to find a  most reachable set $\{(s, t)\}_{t\in\mathbb{T}}$ along the  spatiotemporal dimensions. This reachable set is closed, allowing us to reduce the search by looking only within the enclosed spatiotemporal space.

The above procedure actually views all $H^u_{s_0,s}$ independently, and does not compute the probability for every trajectory. If one state-action pair in a trajectory is removed, the associated trajectory will become invalid in the reduced space. 
Therefore, it may eliminate too many potential trajectories (paths) $\tau$ with relatively large probability.
In order to remedy this problem, 
we reconstruct a TVMDP on the reduced spatiotemporal space to maintain a certain correlation between connected spatial states (recall the definition of connection in section \ref{sec:PPT}). 
In Section~\ref{sec:final-algorithm},
we will see this procedure is equivalent to \textit{mapping} a removed potential trajectory to another one in the reduced search space. 

The final algorithm iterates the above procedures on the whole spatiotemporal space. 
In each iterative step, the first and second moments of the PPT are recomputed.
Intuitively, {\em the first moment is used to find the ``backbone" that outlines the most reachable states, whereas the second moment determines the ``thickness" (or volume) of the most reachable space.}
The policy is then updated on the reduced space, and the actions for states outside the reduced space are mapped to the updated actions on nearest states in the reduced space.
Comparing with~\cite{liu2018solution} which only uses the first moment of PPT to characterize TVMDP,
our method obtains a better approximation to the full spatiotemporal space of TVMDP.

\subsection{Value Iteration with Expected Passage Percolation Time}\label{sec:expected_time}
{
\begin{figure}[t]
    \centering
    \subfigure[]{\label{fig:sigma_0.1}\includegraphics[width=0.45\textwidth]{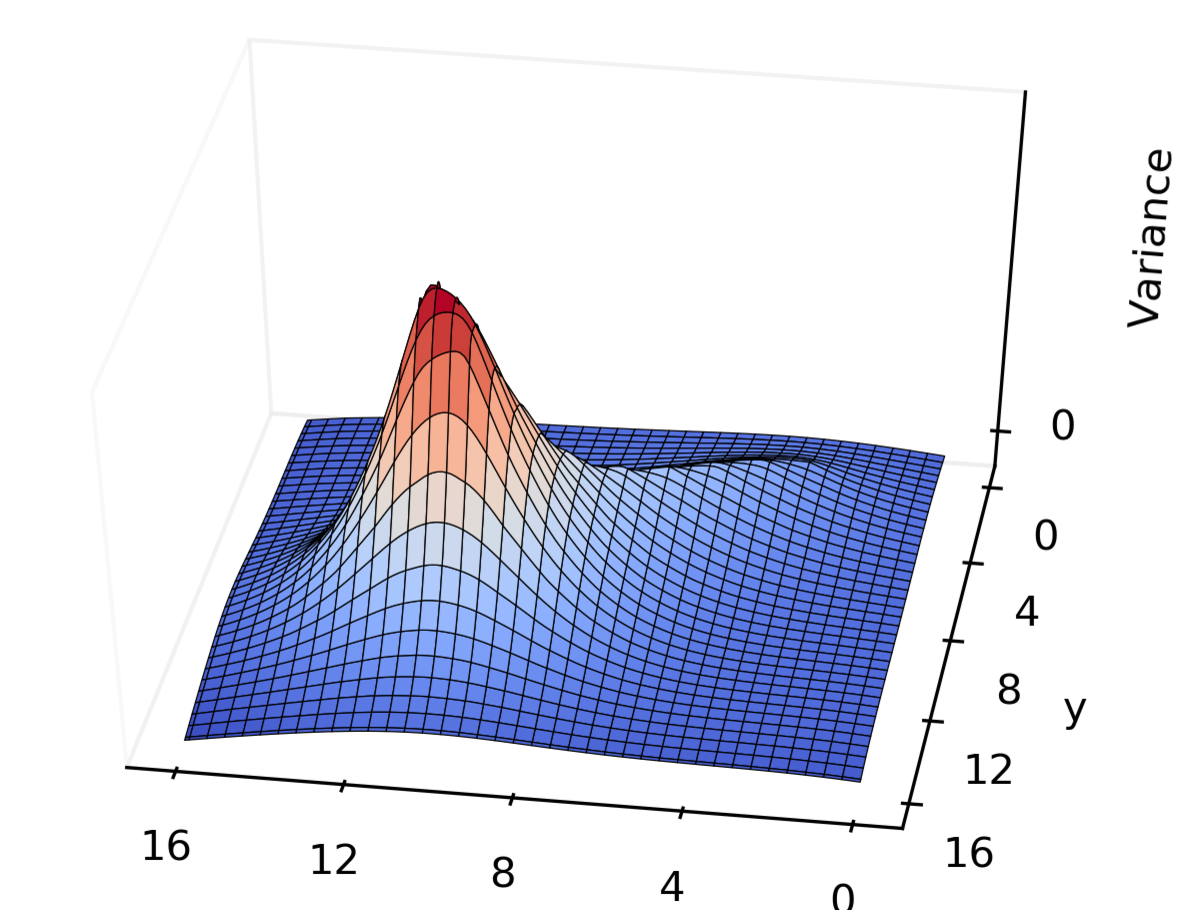}}
    \subfigure[]{\label{fig:sigma_0.1_topdown}\includegraphics[width=0.49\textwidth]{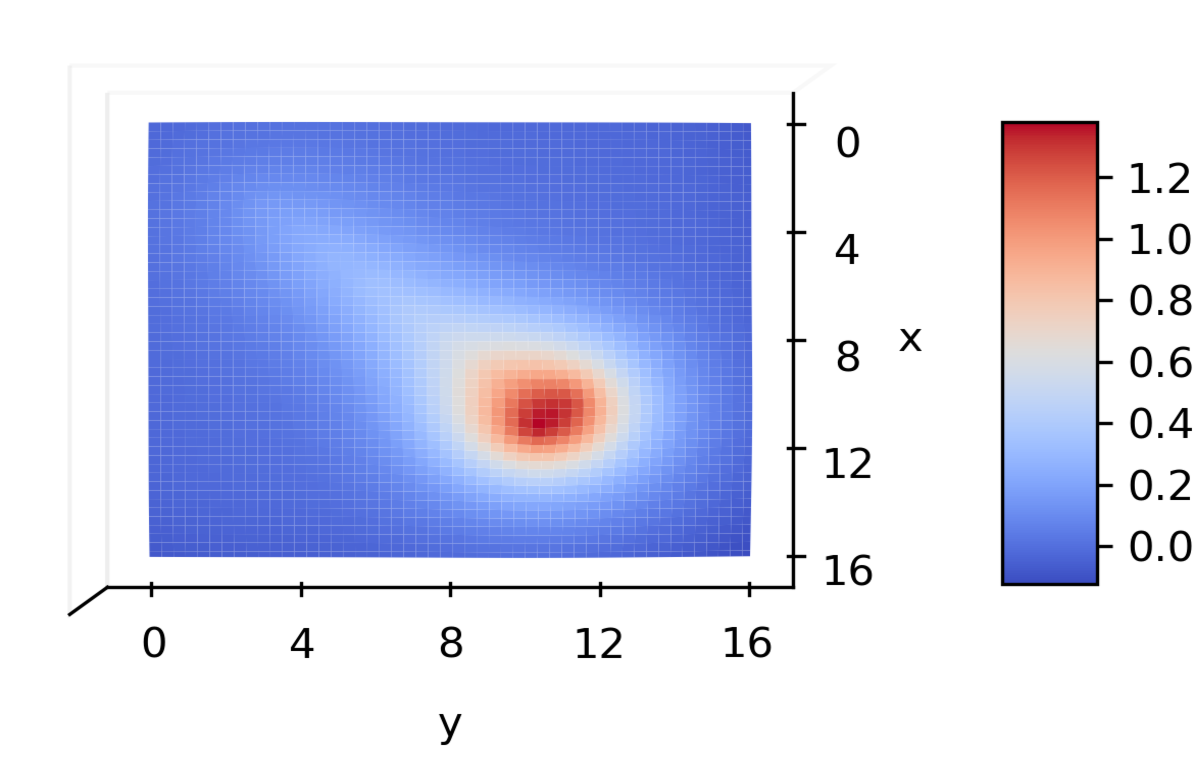}}\\
    \vspace{-10pt}
    \subfigure[]{\label{fig:sigma_1}\includegraphics[width=0.45\textwidth]{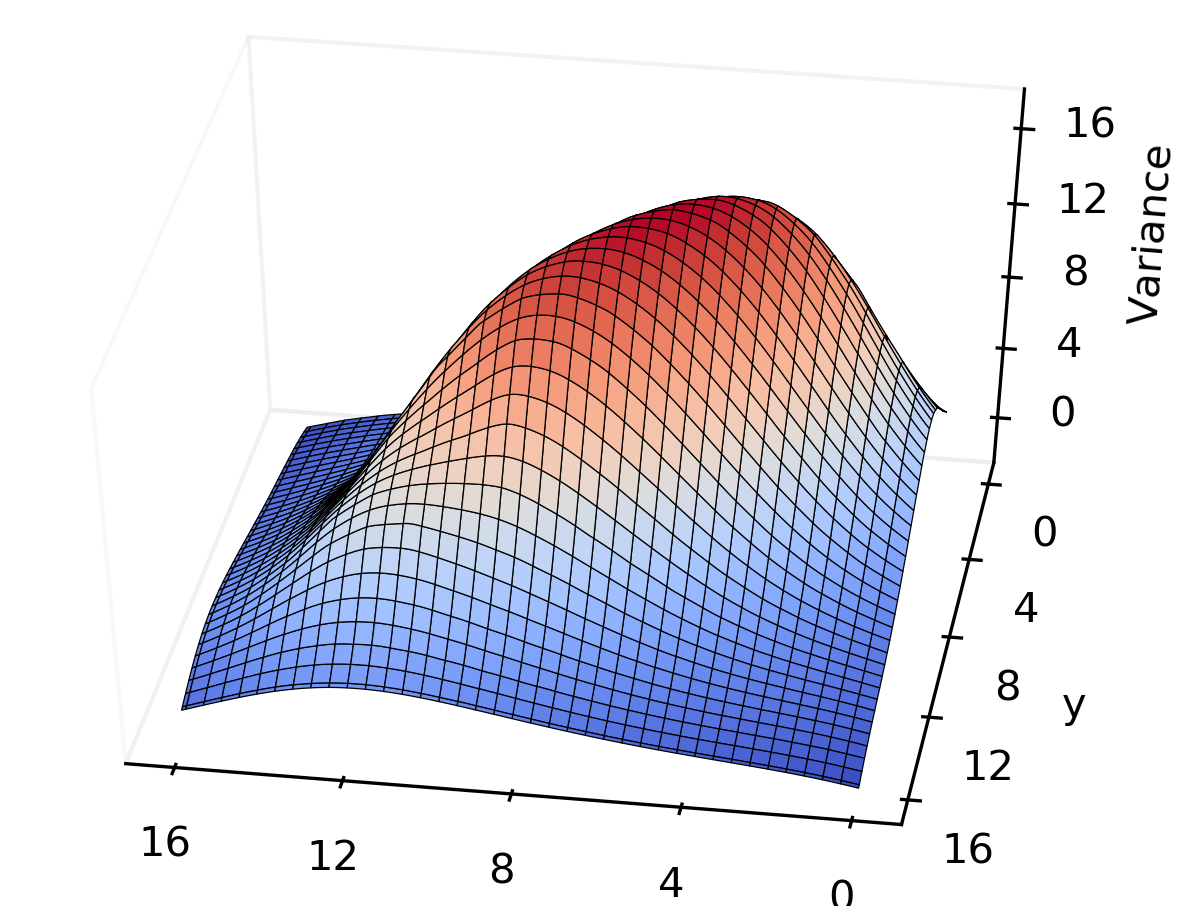}}
    \vspace{-10pt}
    \subfigure[]{\label{fig:sigma_1_topdown}\includegraphics[width=0.49\textwidth]{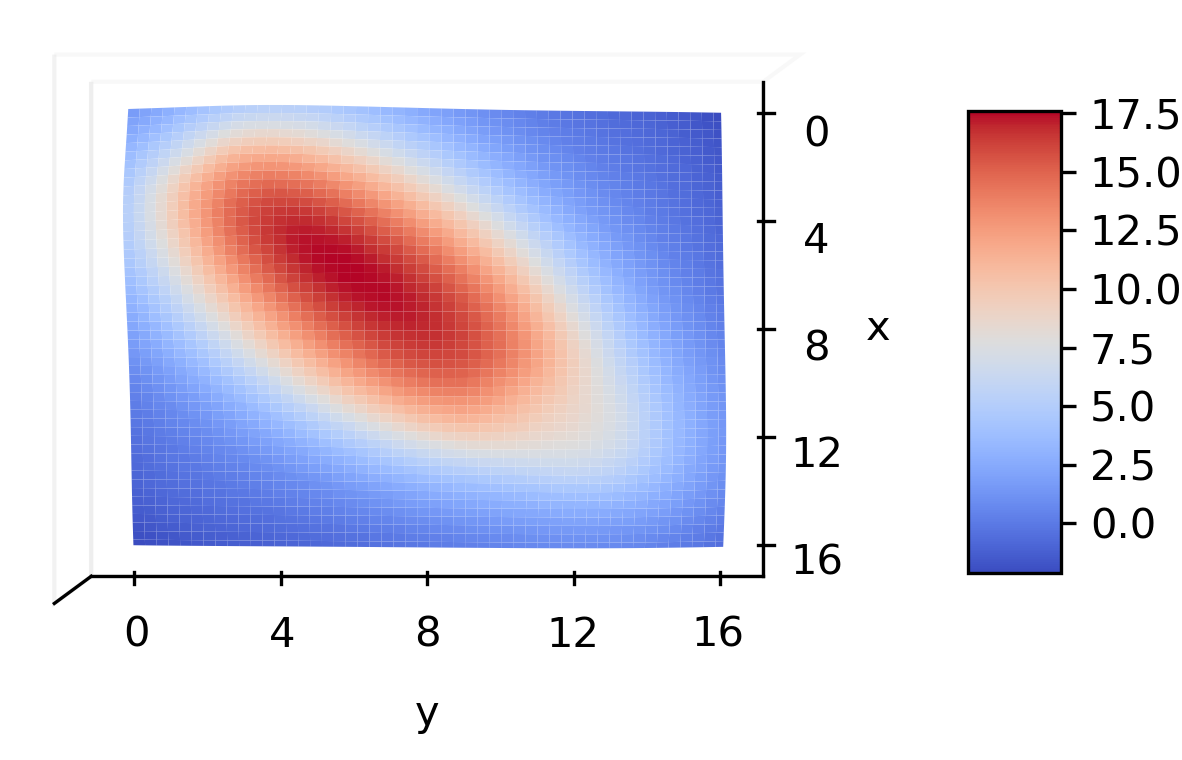}}
    \caption{\small The change in the variance under different uncertainties in the environment. 
    The state transition is modeled as a Gaussian distribution which is referred to Sec.~\ref{sec:experiment}.
    (a)(b) 3D and top down views with the variance of the transition function set to $0.1$.
    (c)(d) A more uncertain environment with the variance set to $1$.        
    }
     \label{fig:variance_}
\end{figure}
}
This section presents the first approximate algorithm, an expected PPT-based value iteration, 
for TVMDPs. It serves as a burn-in procedure for the algorithm in Section \ref{sec:final-algorithm}.


{In this first approximate algorithm, 
we use the transition probabilities and the reward function at the expected PPTs to approximate their time-varying counterpart.}
Then we approximate the TVMDP by an MDP with a properly defined action set at $s$. Accordingly, we can conduct value iteration on the state space $\mathbb{S}$ rather than space $\mathbb{S}\times\mathbb{T}$, and use the resulting time-independent policy for the state at any time. 
 
Suppose $\mu_{s}:=\mathbb{E}^\pi[H^{t_0}_{s_0, s}]$ are already obtained for all $s \in \mathbb{S}$ given the initial state $s_0$ and the starting time $t_0$.
Note that we always have $\mu_{s_0}=0$. With a new transition probability law defined as $\mathcal{T}^{\pi,\mu_s}_{s,s'}:= \sum_{t'}\mathcal{T}(s, \mu_s, a; s', t')$ and a new reward function as $R^{\mu_s}(s, a):=R(s, a, \mu_{s})$, one can update the policy by solving the following Bellman equation
\begin{equation}
    v(s) = \max_{a\in A(s)} \left\{ R^{\mu_s}(s, a) + \gamma\cdot\mathbb{E}^{\pi}_{\mu_s}[v(s')\mid s, a]\right\}, \label{ST-VI-E-PPT}
\end{equation}
where $\mathbb{E}^{\pi}_{\mu_s}(\cdot)$ indicates the calculation under the new transition law $\mathcal{T}^{\pi,\mu_s}_{s,s'}$ and $A(s)$ is the action set at $s$. 
The output policy does not depend on time anymore, and will be used for any state all the time. 
Usually we only have one $t'\in\mathbb{T}$ such that $\mathcal{T}(s, u, a; s', t') >0$ for fixed $s, u, a$ and $s'$ in the practical modeling framework. If $t''\neq t'$, then $\mathcal{T}(s, u, a; s', t'') =0$. In this case, $\mathcal{T}^{\pi,\mu_s}_{s,s'}=\mathcal{T}(s, \mu_s, a; s', t')$.
Now we show how to obtain the expected PPTs. By the conditional expectation formula $\mathbb{E}^{\pi}[H_{s, s'}^u] = \mathbb{E}^{\pi}[H_{s'', s'}^u\mid s'']$,  the expected PPTs satisfy the following linear system
\begin{subequations}\label{eq:mppt}
{\small
\begin{align}
   \mathbb{E}^{\pi}[H_{s,s'}^{u}]&= \sum_{s'', u''} \mathcal{T}(s, a, u; s', u'')\left(h(s, u, s'') + \mathbb{E}^{\pi}[H_{s'',s'}^{u''}]\right) \label{eq:mppt-1}\\
   \mathbb{E}^{\pi}[H_{s,s}^{u}] &=  0 \label{eq:mppt-2},
\end{align}
}
\end{subequations}
where $s\neq s'$, 
$u''=u+h(s, u, s'')$, and $s''$ in Eq.~\eqref{eq:mppt-1} is the next visiting state from $s$. 
The above recursive relationship is similar to the equations for estimating the mean first passage time for a Markov Chain \cite{Jeffrey2018,debnath2018solving}, 
except that the mean first passage time from a state to itself is a positive value.

We do not solve 
Eq.~\eqref{eq:mppt} for all time $u$ altogether. 
Instead, we approximate the solutions in a recursive fashion with the aim of estimating 
$\mu_s$ for all $s$ in the space. In other words, we only carry the estimates of $\mu_s$ into the next iteration. 
In each iteration, we approximate Eq. (\ref{eq:mppt-1}) by the following linear system
\begin{align}
   \mathbb{E}^{\pi}[H_{s,s'}^{\tilde{\mu}_s}] &= \sum_{s''} \mathcal{T}^{\pi,\tilde{\mu}_s}_{s,s''}\left(h(s, \tilde{\mu}_s, s'') + \mathbb{E}^{\pi}[H_{s'',s'}^{\tilde{\mu}_{s''}}]\right), \label{approx-eq:mppt-1}
\end{align}
where $\tilde{\mu}_s=\mathbb{E}^{\tilde{\pi}}(H^{t_0}_{s_0, s})$ are obtained from the latest iteration with the previous policy $\tilde{\pi}$. 
It should be noted that one only needs to solve a linear system which consists of $(N-1)$ linear equations with $N-1$ unknown variables for one $\mu_s$. Here, $N:=|\mathbb{S}|$ denote the total number of states.
Therefore, we merely need to solve $(N-1)$ linear systems to obtain all 
$\mu_s$ in one iteration step. 

\begin{subfigures}
\begin{figure}
\end{figure}
\begin{algorithm} [t]
    \caption{Value Iteration with Expected Passage Percolation Time }\label{alg:vi_mfpt} 
  \begin{algorithmic}[1]
     \INPUT{Time-varying transition function $\mathcal{T}$, time-varying reward $R$, spatial states $\mathbb{S}$, and $t_0=0$.}
    \OUTPUT{ Policy $\pi$ and the expected passage percolation time $\mu_s, \forall s \in \mathbb{S}$.}
    \State Initialize $\mu_s=\tilde{\mu}_s=0, v(s)=0, \forall s \in \mathbb{S}$;
    \Repeat
      \State $\delta = 0$.
      \ForEach {$s \in S$}
          \State $V = v(s)$;
          \State Solve Bellman equation (\ref{ST-VI-E-PPT});
          \State $\delta = \max(\delta, V - v(s))$;
      \EndFor
      
      \State \parbox[t]{.8\linewidth}{Get policy $\pi$;}
      \State \parbox[t]{.8\linewidth}{Solve for $\mu_s$ by $N-1$ systems of linear equations (\ref{approx-eq:mppt-1});
      }
      \State \parbox[t]{.8\linewidth}{Set $\tilde{\mu}_s = \mu_s$;}
      
    \Until $\delta$ is less than a threshold or the maximum number of iteration has been reached. 
  \end{algorithmic}
\end{algorithm}
\end{subfigures}

Lastly, we make a note on the numerical solution for solving the above linear system. 
In practice, there could be extreme large values for some estimates of $\mathbb{E}(H^{\tilde{\mu_s}}_{s,s'})$, 
indicating that the associated states are nearly impossible to visit. 
Yet these large values could result in numerical instability. 
To avoid this issue, we put a discount factor $\alpha$ $<$1.0 to the expected PPT on the right side of Eq. (\ref{approx-eq:mppt-1}), 
i.e., replacing $\mathbb{E}^{\pi}[H_{s'',s'}^{\tilde{\mu}_{s''}}]$ by $\alpha\cdot\mathbb{E}^{\pi}[H_{s'',s'}^{\tilde{\mu}_{s''}}]$.
%
The expected PPT-based algorithm is described in Alg. \ref{alg:vi_mfpt}.

\subsection{Variance of Passage Percolation Time and Reachable Space}\label{sec:sec_moment}
Once the expectation of $H_{s, s'}^{u}$ is obtained, 
one is able to further derive the estimation of variance under the same policy $\pi$ by the total variance formula as below \cite{grimmett2001probability}:
\begin{align}
\mathbb{V}^{\pi}[H_{s, s'}^u] &= \mathbb{E}^{\pi}[\mathbb{V}^{\pi}[H_{s, s'}^{u}|s'']] + \mathbb{V}^{\pi}[\mathbb{E}^{\pi}[H_{s, s'}^{u}|s'']],
\end{align} 
where $s''$ can be viewed as the next state to visit from $s$ before reaching $s'$. The two terms on the right hand side of the above equation are calculated as
\begin{align}\label{exp-var-ppt}
\mathbb{E}^{\pi}[\mathbb{V}^{\pi}[H_{s, s'}^{u}|s'']] &=\sum_{s'', \mu''}\mathcal{T}(s, u, a; s'', \mu'')\mathbb{V}^{\pi}[H_{s'', s'}^{u''}],
\end{align} 
and
\begin{eqnarray}\label{var-exp-ppt}
&{}&\mathbb{V}^{\pi}[\mathbb{E}^{\pi}[H_{s, s'}^{u}|s'']]\nonumber\\
&=& \sum_{s'', u''}\mathcal{T}(s, u, a; s'', u'')(\mathbb{E}^{\pi}[H_{s, s'}^{u}|s''] - \mathbb{E}^{\pi}[H_{s, s'}^u])^2\nonumber\\
&=& \sum_{s'', u''}\mathcal{T}(s, u, a; s'', u'')\cdot\left(h(s, u, s'')\right. \nonumber\\
&{}&\left.+ ~\mathbb{E}^{\pi}[H_{s'', s'}^{u''}] - \mathbb{E}^{\pi}[H_{s, s'}^u]\right)^2.
\end{eqnarray}
Eq. (\ref{exp-var-ppt}) and (\ref{var-exp-ppt}) show that the variance estimation also relies on solving a linear system. 
Because the local transition time $h(s, u, s')$ is assumed deterministic and does not depend on policy, 
it can be taken out from the term $\mathbb{E}^{\pi}[H_{s, s'}^{u}|s'']=\mathbb{E}^{\pi}[h(s, u, s'')+H_{s'', s'}^{u}|s'']$ in Eq. (\ref{var-exp-ppt}). The same idea also applies to $\mathbb{V}^{\pi}[H_{s, s'}^{u}|s'']$ in Eq. (\ref{exp-var-ppt}). 

Because we aim to obtain a range of time around $\mu_s$ to reach state $s$ from the initial state $t_0$, 
we only focus on $\sigma_s^2:=\mathbb{V}^{\pi}[H_{s_0, s}^{t_0}]$. 
The structure of linear equations for $\sigma_s^2$ is similar to that for expectation in the previous section. 
Therefore, we can get the solutions in a similar recursive fashion:
\begin{align}\label{eq:variance}
 &\mathbb{V}^{\pi}[H_{s, s'}^{\mu_s}]  = \sum_{s''}\mathcal{T}_{s,s''}^{\pi, \mu_s}\left(\mathbb{V}^{\pi}[H_{s'',s'}^{\mu_s}] \right. \nonumber\\
 &\left.+ \left(h(s, \mu_s, s'')+\mathbb{E}^\pi[H_{s'', s'}^{\mu_{s''}}] - \mathbb{E}^{\pi}[H_{s, s'}^{\mu_{s}}]\right)^2\right).
\end{align}

Similarly, an array of linear $N-1$ equations~\eqref{eq:variance} with the same number of unknown variables needs to be solved for a $\sigma_s^2$. 
It is worth noting that the computation of expectation and variance of PPT can be computed in parallel, 
offering an extra boost in computational time as shown later in Section \ref{sec:experiment}. 
As an illustration, Fig. \ref{fig:variance_} shows the variance of PPT from the initial state to other states in spatial 
(width of top down view) and temporal (height of the 3D view) space under different environmental uncertainties.

Now we can formally define the reachable space of TVMDPs. 
\begin{definition}\label{def:reachable_states}
Given a policy $\pi$ for TVMDP, a state-time pair $(s, t)$ is reachable from an initial state-time pair $(s_0, t_0)$ if $t$ satisfies
\begin{equation}\label{eq:reachable_states}
    \mu_s - m_r\sigma_s \leq t \leq \mu_s + m_r\sigma_s,
\end{equation}
where $m_r \geq 1$ is a predefined regulation parameter.
The set of all reachable state-time pairs is called reachable space for TVMDP, and is denoted by $(\mathbb{S}\times \mathbb{T})^{-}$.
\end{definition}
Note that the predefined parameter $m_r$ used to further ``inflate" the reachable space envelope if there is a need. 

\subsection{TVMDP Value Iteration with Reachable Space}\label{sec:final-algorithm}
\addtocounter{figure}{-1}
\begin{figure}[t]
    \vspace{3pt}
    \centering
    \includegraphics[width=0.45\textwidth]{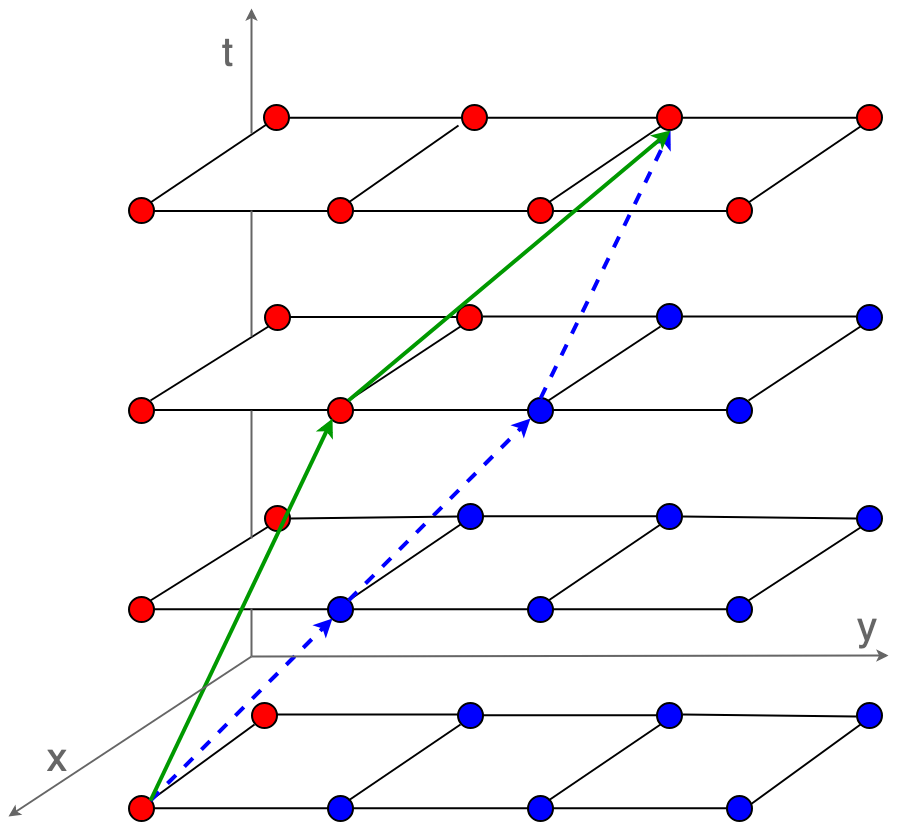}
    \caption{\small Illustration of the reconstruction procedure.
Four replicated layers on different time slots. Transition only between neighbored spatial states in two consecutive time slots. The red nodes are spatiotemporal states in reachable space, whereas blue ones are outside. Transition probability from the bottom left state to the next state (the end of blue arrow) is reassigned to the state two time slots away (the end of green arrow).
}
\label{fig:re_route}
\end{figure}
The reachable space describes the variability of the first time for the vehicle to visit a state. 
Computational cost may be reduced by iterating within the reachable space only.
However, the space trimming leads to some issues while processing the states at or around the reachable space border.

Consider a random trajectory $\{\ldots, (s_{k}, a_{k}),(s_{k+1}, a_{k+1}), $ $ \ldots\}$, where the vehicle reaches $s_{k}$ at $t_k$ with $(s_{k}, t_k)$ in the reachable space $(\mathbb{S}\times \mathbb{T})^{-}$. To aid a clear presentation, we 
use the symbol $\mathcal{T}_u$ as a shorthand for $\mathcal{T}(s_k, t_k, a_k; s_{k+1}, u)$ in this subsection.  
If $(s_{k+1}, t_{k+1})$ is not in this reachable space, the transition probability law restricted on $(\mathbb{S}\times \mathbb{T})^{-}$ 
has the property 
$\sum_{u:(s_{k+1},u)\in(\mathbb{S}\times \mathbb{T})^{-}}\mathcal{T}_{u} < 1$. This places a negative impact on 
red the algorithms to maximize the value function.
Therefore, we need to reassign the transition probability law as follows.

Let $l_{s_{k+1}}=\mu_{s_{k+1}} - m_r\cdot\sigma_{s_{k+1}}$ and $b_{s_{k+1}}=\mu_{s_{k+1}} + m_r\cdot\sigma_{s_{k+1}}$ 
If $t_{k+1} < l_{s_{k+1}}$, then we reassign transition probability 
\begin{equation*}
\mathcal{T}_{l_{s_{k+1}}} = \sum_{u'\leq l_{k+1}} \mathcal{T}_{u'}.
\end{equation*}
For $t_{k+1} > l_{s_{k+1}}$, we simply re-normalize the transition probability if $0<\sum_{u:(s_{k+1},u)\in(\mathbb{S}\times \mathbb{T})^{-}}\mathcal{T}<1$. We call such a treatment a \textit{reconstruction} procedure for TVMDPs. 

Fig.~\ref{fig:re_route} illustrates an intuition of the reconstruction procedure: if there are a few states along a spatiotemporal path outside the reachable space, the entire path would not be considered in the value function. However, the reconstruction procedure \textit{maps} this path back to the reachable space. 

Our final algorithm combines the techniques introduced in the previous sections and applies value iteration in the reachable space with reconstruction in an iterative manner. It is summarized in Alg.~\ref{alg:vi_reduced}. In each iteration, there are two stages.
    During the first stage, the expectation and variance of PPTs are computed. These estimations allow us to determine the reachable space $(\mathbb{S}\times\mathbb{T})^{-}$.
During the second stage, the value iteration algorithm is employed on $(\mathbb{S}\times\mathbb{T})^{-}$ to update the policy. 
Finally, if $(s, t)\notin(\mathbb{S}\times\mathbb{T})^{-}$, we assign the action $\pi(s, t)$ as  
\begin{align}\label{map-action}
&\pi(s, t) = \pi(s', t), \nonumber\\
&\mbox{where }s'=\argmin_{s'}\{d(s, s')\mid (s',t)\in(\mathbb{S}\times\mathbb{T})^{-} \},
\end{align}
assuming that $\mathbb{S}$ has a metric measure $d(\cdot, \cdot)$.
\begin{subfigures}
\begin{figure}
\vspace{1pt}
\end{figure}
 \begin{algorithm}[t]
   \caption{TVMDP Value Iteration with Reachable Space}
    \label{alg:vi_reduced}
   \begin{algorithmic}[1]
   \INPUT{TVMDP elements $(\mathbb{S}, \mathbb{T}, \mathbb{A}, \mathcal{T}, R)$; starting state $s_0$; starting time $t_0:=0$; $m_r$; numbers $k, n, i$.}
    \OUTPUT{$\pi^*$: an approximation of the optimal policy $\mathbb{S}\times\mathbb{T}$.}
   
   \State Burn-in: run Algorithm \ref{alg:vi_mfpt} for $k$ iterations to get $\pi_0$;
   \For{\texttt{j=0...n}}
        \State Compute $\mathbb{E}^{\pi_j}[H_{s_0, s}^{t_0}]$ and  $\mathbb{V}^{\pi_j}[H_{s_0, s}^{t_0}]$ for all $s\in\mathbb{S}$;
        \State Apply Eq. \eqref{eq:reachable_states} to obtain reachable space $(\mathbb{S}\times\mathbb{T})^{-}$;
        \State Perform reconstruction procedure;
        \State Execute value iteration for $i$ iterations to obtain $\pi_{j+1}$ on $(\mathbb{S}\times\mathbb{T})^{-}$;
        \State Map actions using Eq.(\ref{map-action}) if $(s, t)\notin(\mathbb{S}\times\mathbb{T})^{-}$;
        \If{$\pi_{j-1} = \pi_{j}$ and $j>0$}
        \State break
        \EndIf
    \EndFor
   \end{algorithmic}
   \label{alg:Variance-PPT-VI}
\end{algorithm}
\end{subfigures}

\section{EXPERIMENTS}\label{sec:experiment}

We validate our proposed method in simulation. The experiments were conducted in Ubuntu 16.04 on a PC with a 4.20 GHz i7-7700k CPU and 32 GB RAM.

\addtocounter{figure}{-1}
\begin{figure}[t]
    \centering
        \vspace{5pt}
    \includegraphics[width=0.6\textwidth]{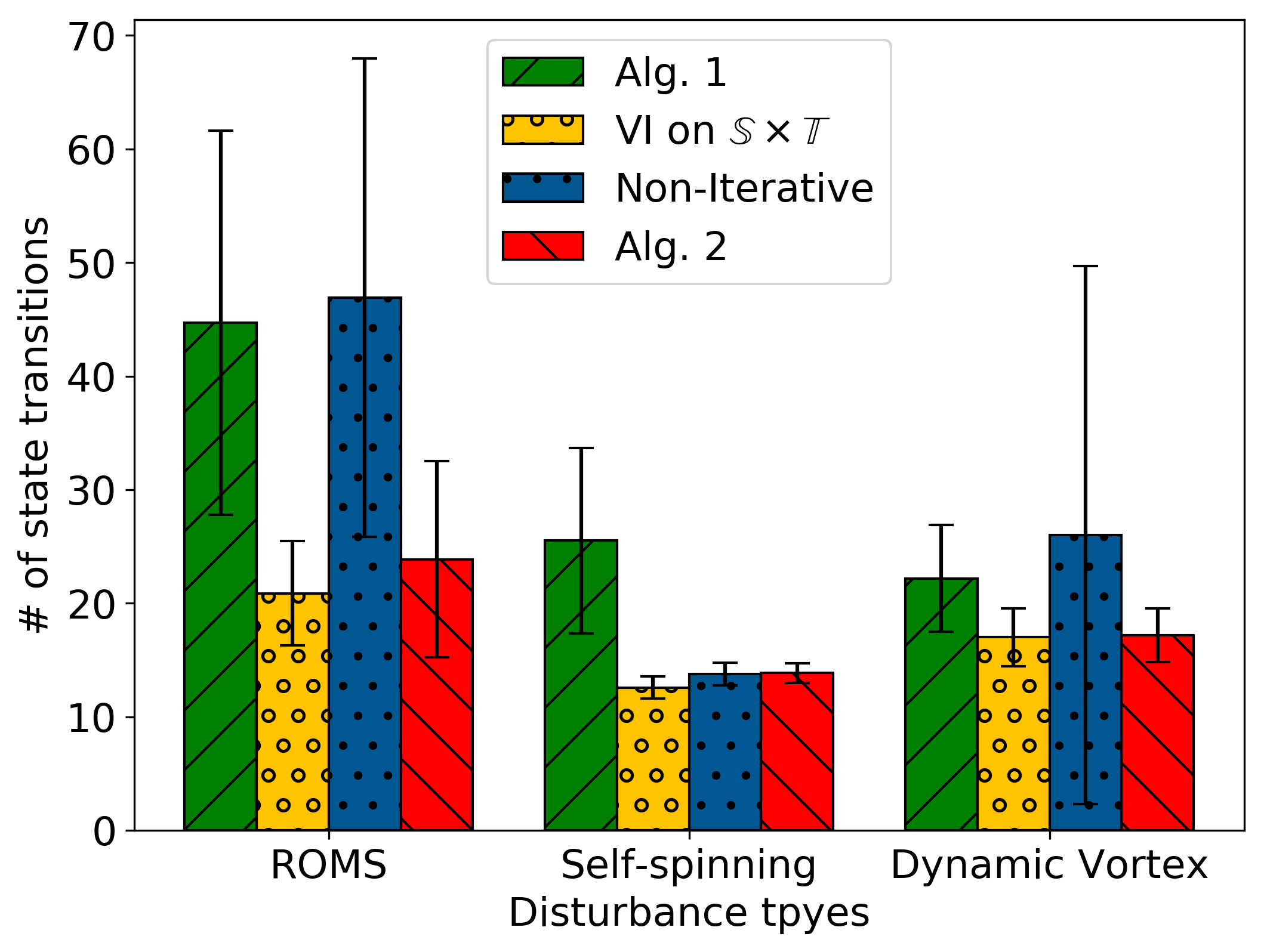}
    \caption{\small Comparison of averaged number of transitions for four algorithms. The result averages over 100 runs with the same random seed for each algorithm and disturbance.}
    \label{fig:toy_exp} 
\end{figure}

\begin{figure}[t]
    \centering
    \includegraphics[width=0.6\textwidth]{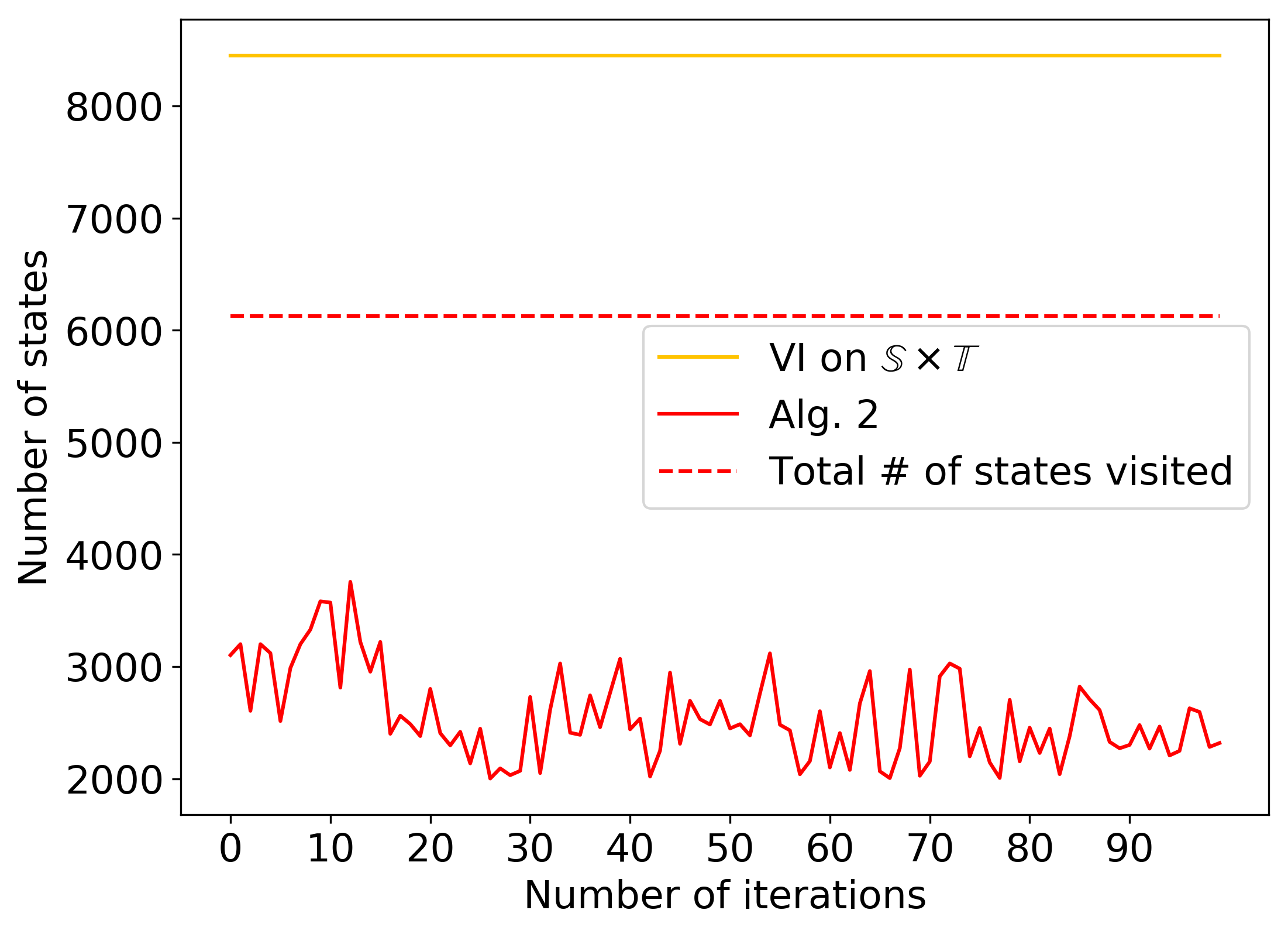}
    \vspace{-5pt}
    \caption{\small Comparison between the number of states in the full space and reduced space over a hundred iterations in $13\times13$ grid world.         \vspace{-15pt}}
    \label{fig:num_states}
\end{figure}

\subsection{Experimental Setup}

\begin{figure*} 
  \centering 
  \subfigure[VI on $\mathbb{S}\times\mathbb{T}$]
        {\label{fig:optimal_spin}\includegraphics[width=0.165\textwidth]{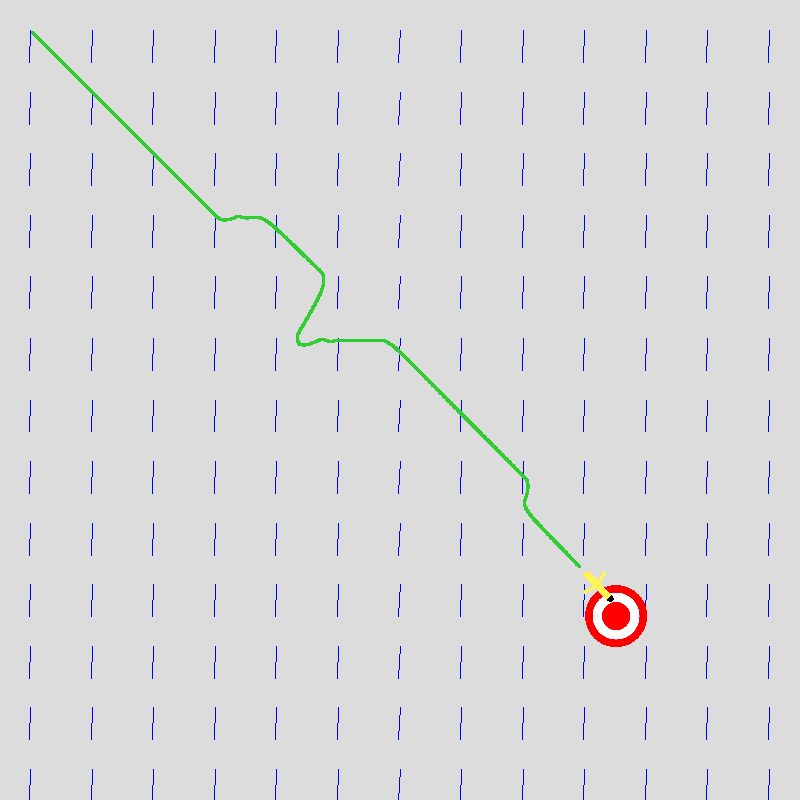}}
  \qquad
  \subfigure[Alg.~\ref{alg:vi_reduced}]
        {\label{fig:iter_spin}\includegraphics[width=0.165\textwidth]{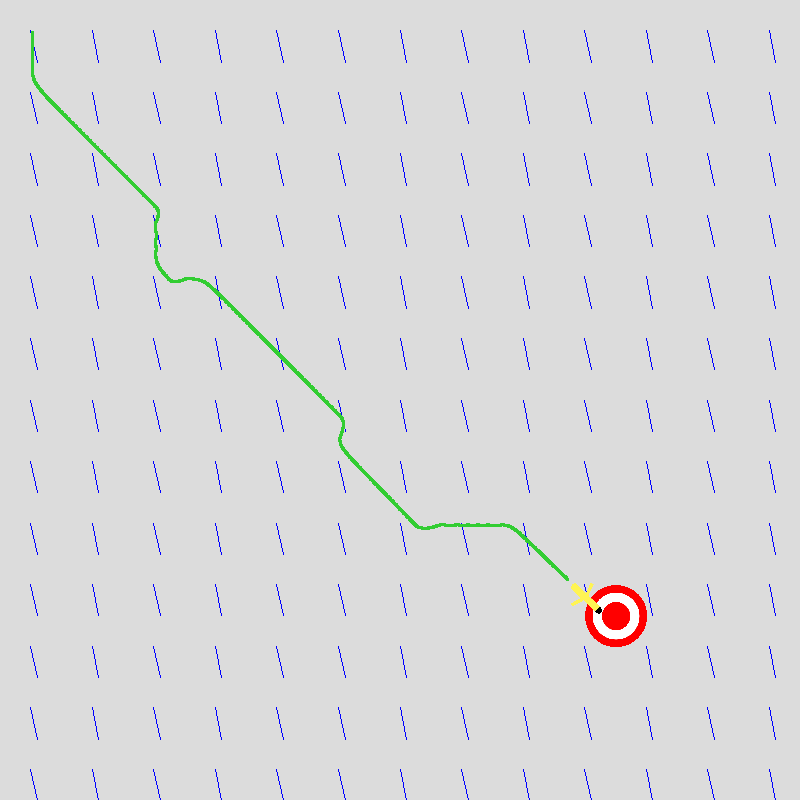}}
  \qquad
  \subfigure[Non-iterative]
        {\label{fig:non_iter_spin}\includegraphics[width=0.165\textwidth]{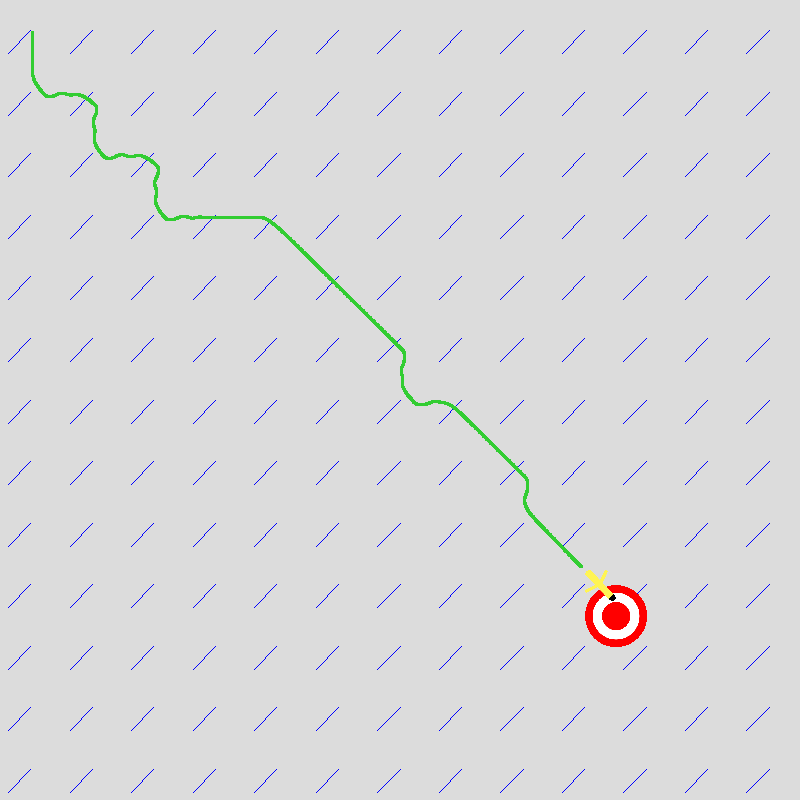}}
  \qquad
  \subfigure[Alg.~\ref{alg:vi_mfpt}]
        {\label{fig:tvmdp_spin}\includegraphics[width=0.165\textwidth]{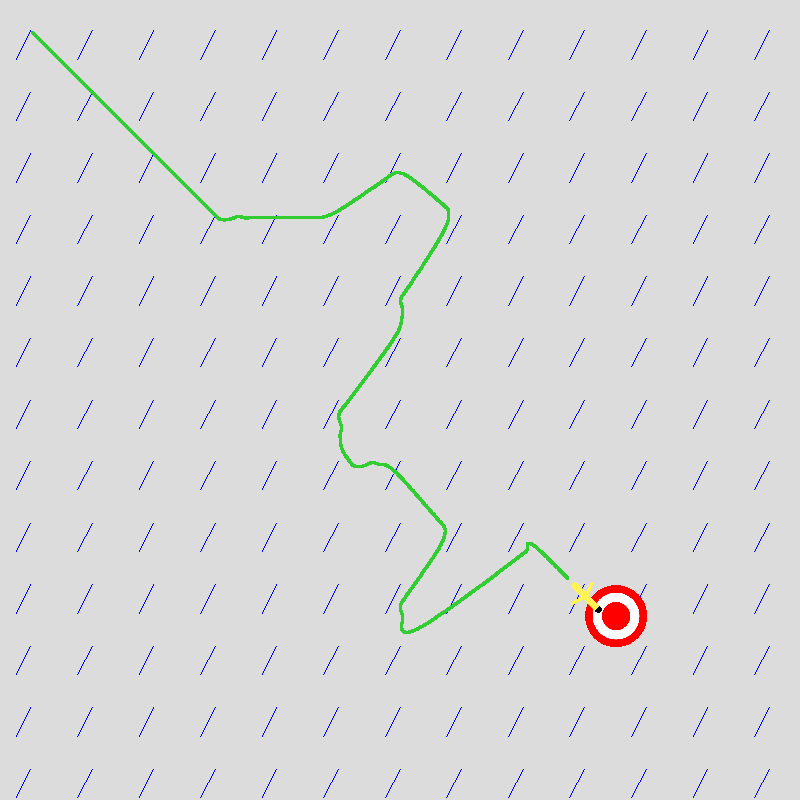}}
  \qquad
  \subfigure[VI on $\mathbb{S}\times\mathbb{T}$]
        {\label{fig:optimal_vortex}\includegraphics[width=0.165\textwidth]{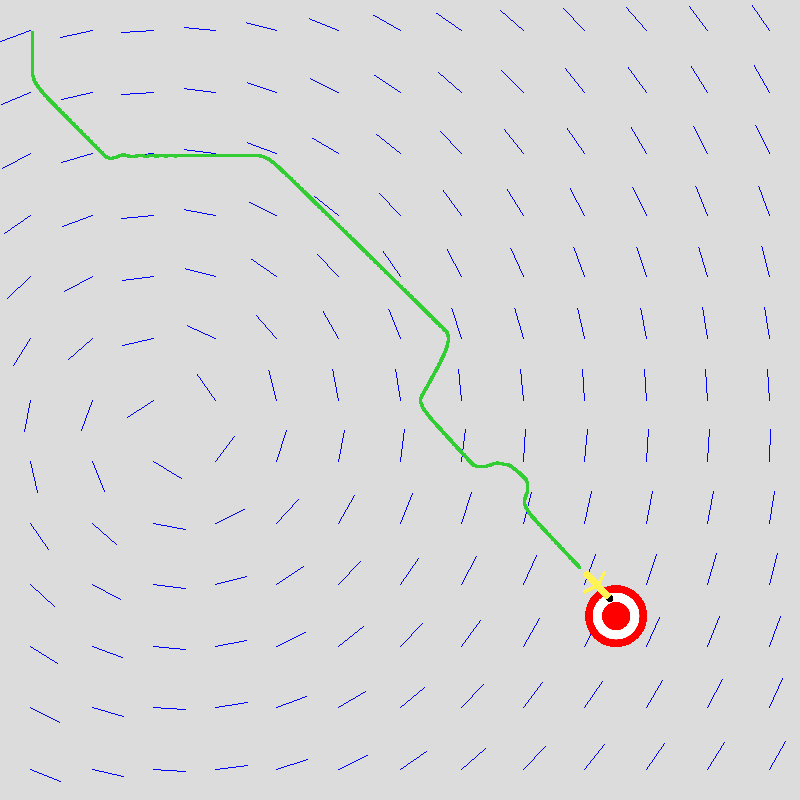}}
  \qquad
   {
       \subfigure[Alg.~\ref{alg:vi_reduced}]
            {\label{fig:iter_vortex}\includegraphics[width=0.165\textwidth]{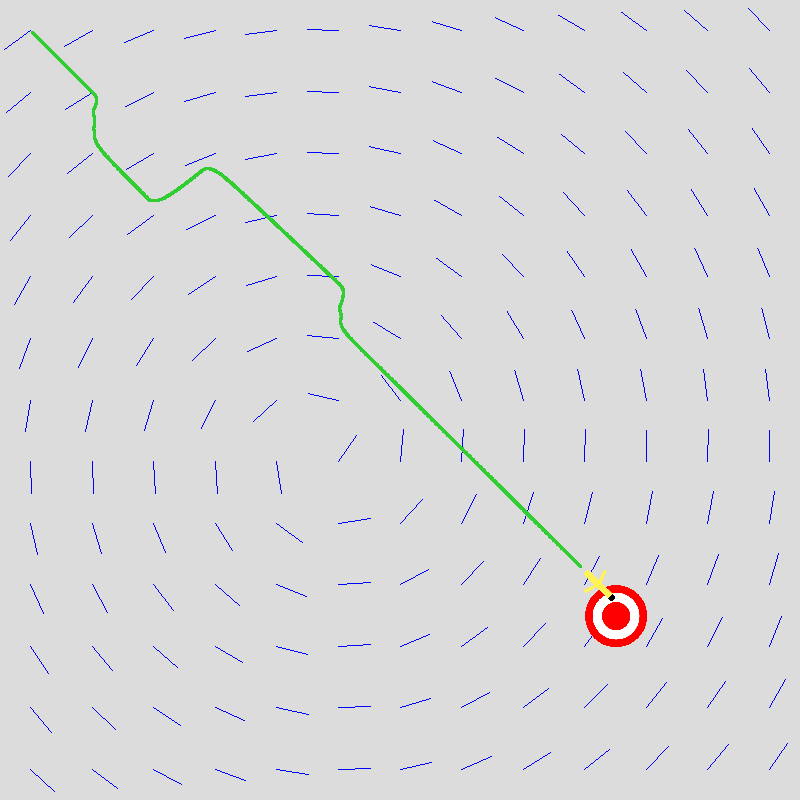}}
        }
  \qquad
  \subfigure[Non-iterative]
        {\label{fig:non_iter_vortex}\includegraphics[width=0.165\textwidth]{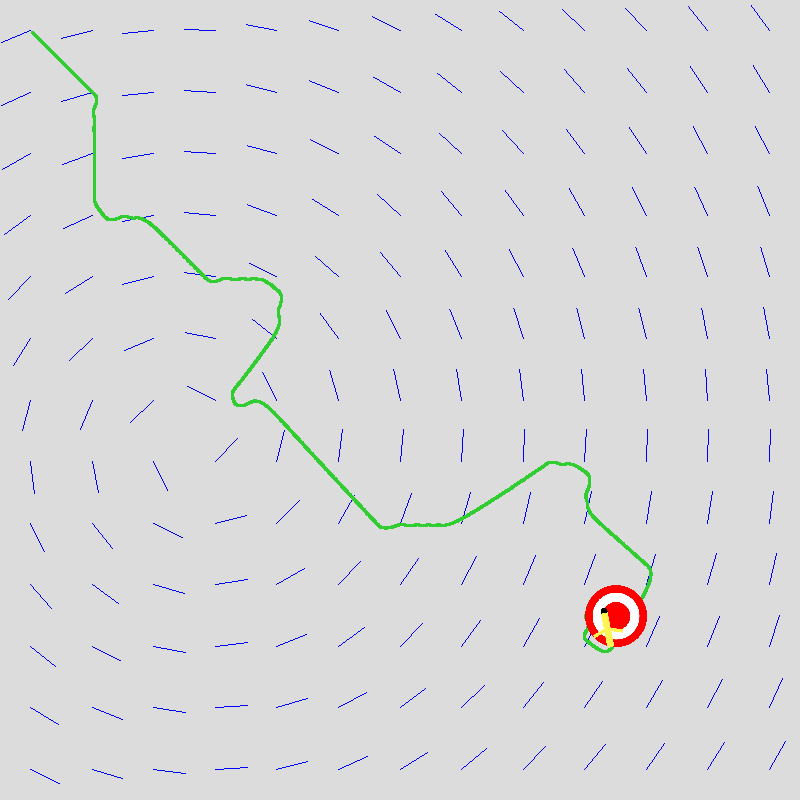}}
  \qquad
  \subfigure[Alg.~\ref{alg:vi_mfpt}]
        {\label{fig:tvmdp_vortex}\includegraphics[width=0.165\textwidth]{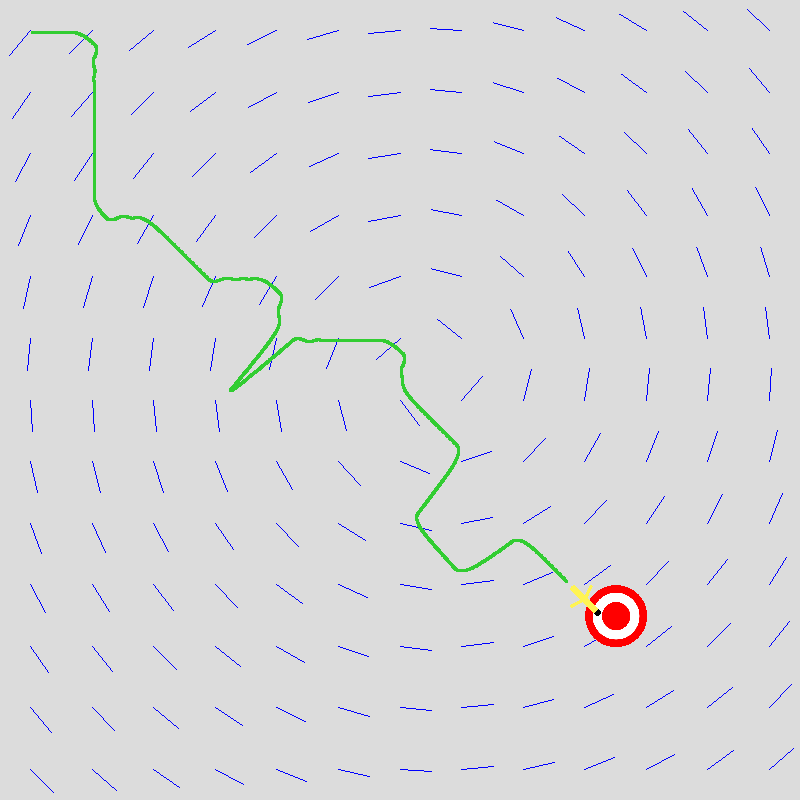}}
  \caption{\small Trajectories in the simplified environment with artificial disturbances. The top four figures demonstrate the trajectories of the four algorithms under self-spinning ocean currents. The bottom row shows the trials under dynamic vortex disturbances.}
\label{fig:toy_traj} \vspace{-10pt}
\end{figure*}

We consider 
a 2D ocean surface subject to time-varying ocean currents in our experiments.
The surface is tessellated into a 2D grid map where the centroid of each cell represents a state. 
A simulated marine vehicle with a kinematics model can transit in eight directions 
$\{N, NE, E, SE, S, SW, W, NW\}$ except for boundary states. 
The task for the vehicle is to reach a designated goal state with the minimum trajectory length. 

Ocean currents are modeled as a velocity vector field where each vector can represent the magnitude and the direction of currents. Three types of currents are considered:
\begin{enumerate}
    \item Self-spinning: each disturbance vector spins at the center of each cell. 
    The vector components are given by $v_x = A\text{cos}(\omega t), v_y = A\text{sin}(\omega t)$, where $A$ is the magnitude and $\omega$ is the spinning  frequency. 
    \item Dynamic vortex: the vector components along $x$-axis and $y$-axis 
    are given by $v_x = x_c - x + y - y_c$, $v_y = x_c - x - y + y_c$, where $(x,y)$ and $(x_c,y_c)$ are coordinates of the cell and the center of the vortex,
    respectively. The center of the vortex translates and rotates according to the following functions $x_c = r\text{cos}(\omega t) + c_x$, $y_c = r\text{sin}(\omega t) + c_y$, where $r$ is the rotating radius and $(c_x, c_y)$ represents the rotating center.
    \item Regional Ocean Model System (ROMS) data \cite{shchepetkin2005regional}: 
    the dataset provides ocean currents data every three hours per day. 
    This allows us to obtain the transition function in the temporal dimension. 
    Gaussian Process Regression (GPR) is used to model 
    and extrapolate
    the ocean current vector field in time and space.
\end{enumerate}

We compare our algorithm (Alg.~\ref{alg:vi_reduced}) 
with three other methods:
\begin{itemize}
    \item Value Iteration in the spatiotemporal space (VI in $\mathbb{S} \times \mathbb{T}$): this method computes solutions exhaustively in the entire space, thus it is used as the baseline (only for problems with a small size state space).
    \item Value iteration with expected PPT (Alg.~\ref{alg:vi_mfpt}): it executes VI in the spatial space $\mathbb{S}$ only, where the time input for transition function is approximated by expected PPTs, i.e., expected PPTs are merely used in Bellman equation (\ref{ST-VI-E-PPT}). This is also the main framework proposed in~\cite{liu2018solution}.
    \item Non-iterative variant of Alg.~\ref{alg:vi_reduced}: it only reconstructs reachable space once (with $n=1$). 
    VI is executed only in this reduced space until the convergence condition is reached. This is used to show the necessity of the iterative procedure in Alg.~\ref{alg:vi_reduced}.
\end{itemize}
We choose $m_r=2$ in Eq.\eqref{eq:reachable_states}.
Because we are interested in minimizing the travel time, we impose a small reward $-0.1$ for each action execution except for reaching the goal and obstacle states.
The rewards for reaching the goal and obstacle states are $1$ and $-1$, respectively.
Note that we do not use time-varying reward functions for experiments, but similar solutions can be obtained by exactly the same algorithm.

\subsection{Simplified and Analytic Scenarios}

\begin{figure}[t]
  \centering 
  \vspace{-15pt}
  \subfigure[]
        {\label{fig:single_thread}\includegraphics[width=0.48\textwidth]{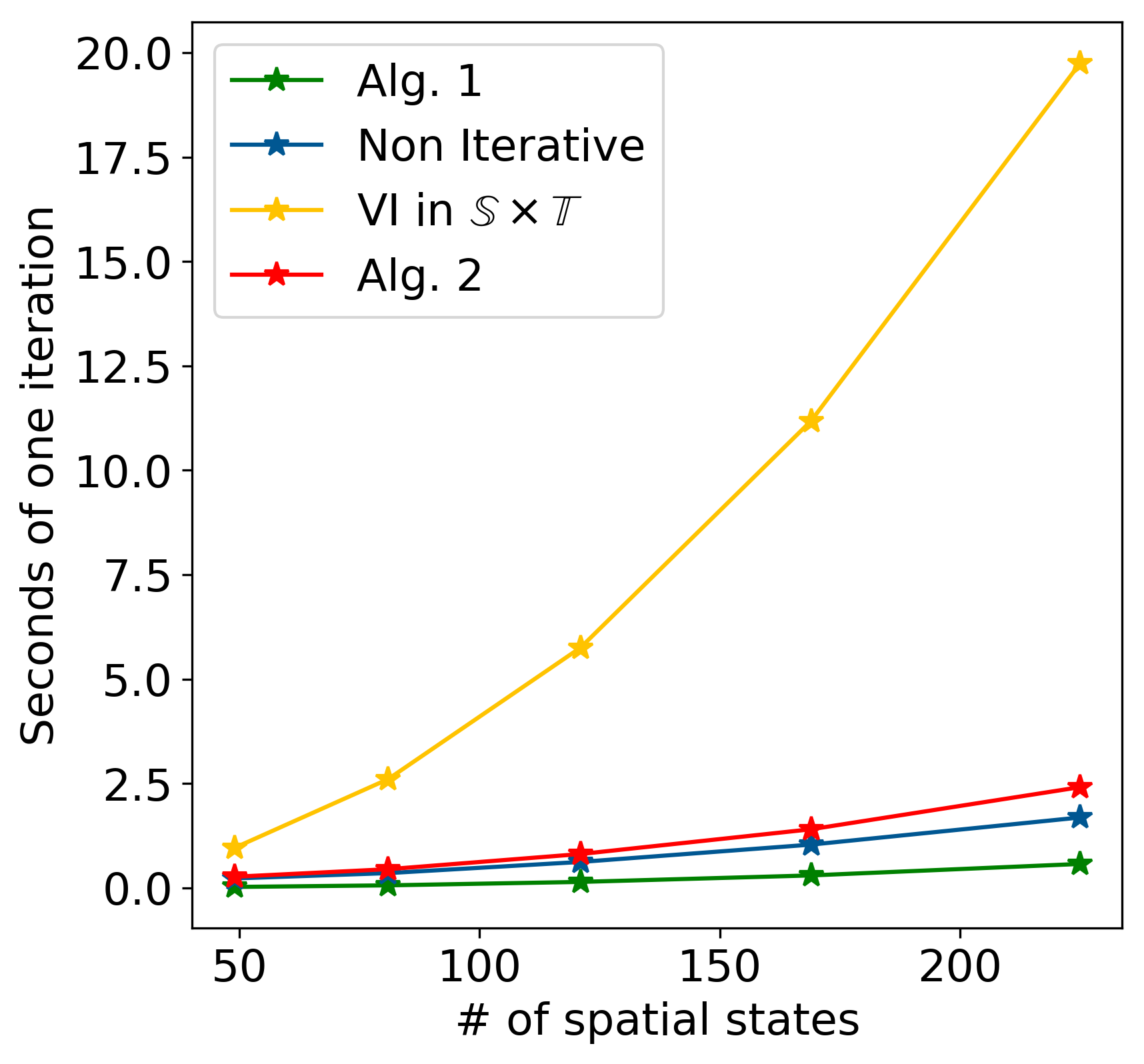}}
  \subfigure[]
        {\label{fig:multi_thread}\includegraphics[width=0.48\textwidth]{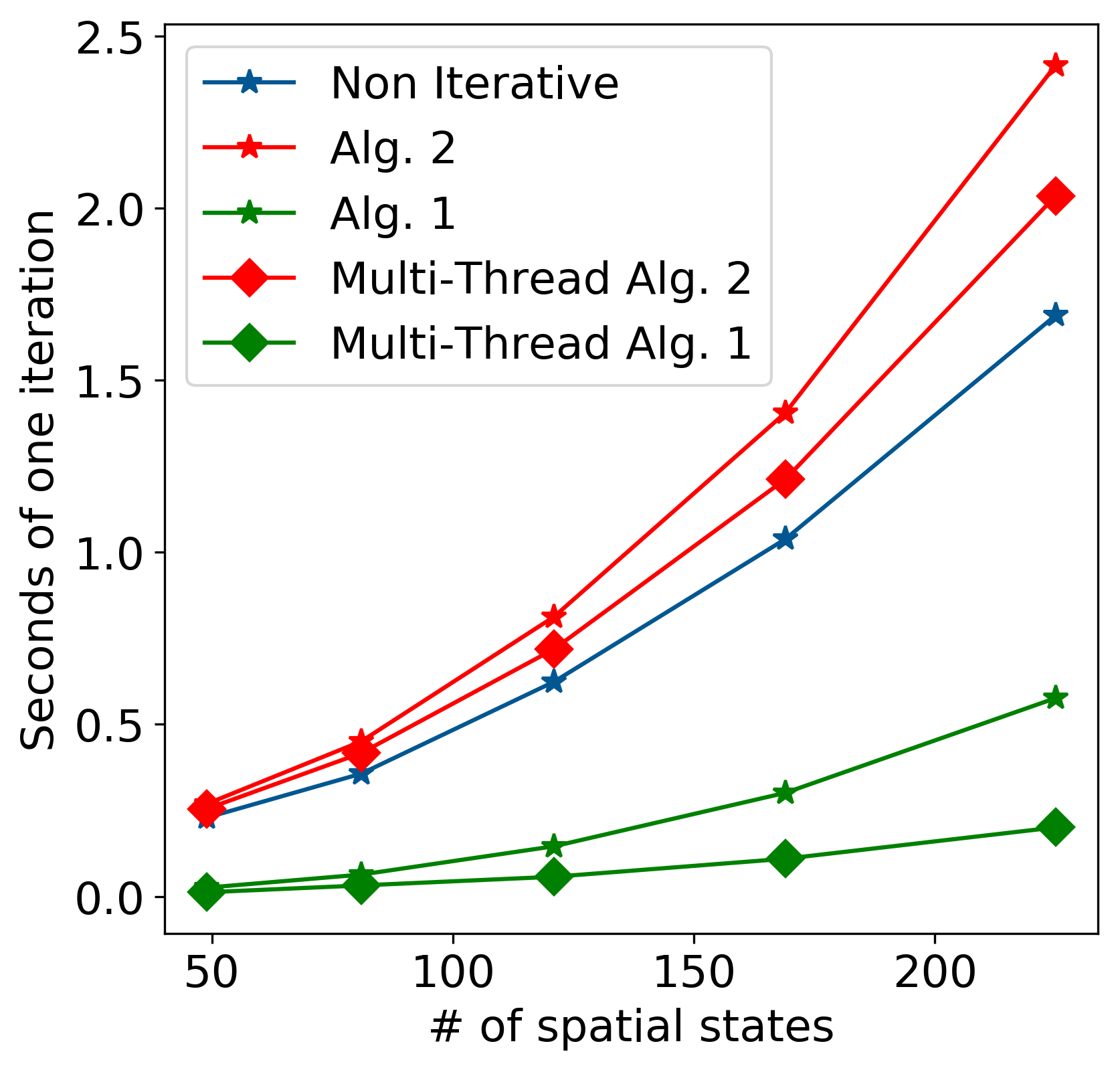} } \vspace{-15pt}
  \caption{\small Computational time comparisons between the four algorithms; (a) Single-threaded version; (b) Multi-threaded version for solving expected times and their variances.        
  }
\label{fig:time_complexity}
\end{figure}

\begin{figure*}[t]
    \centering
    \subfigure[]{\label{fig:full_state_real}\includegraphics[width=0.2\textwidth]{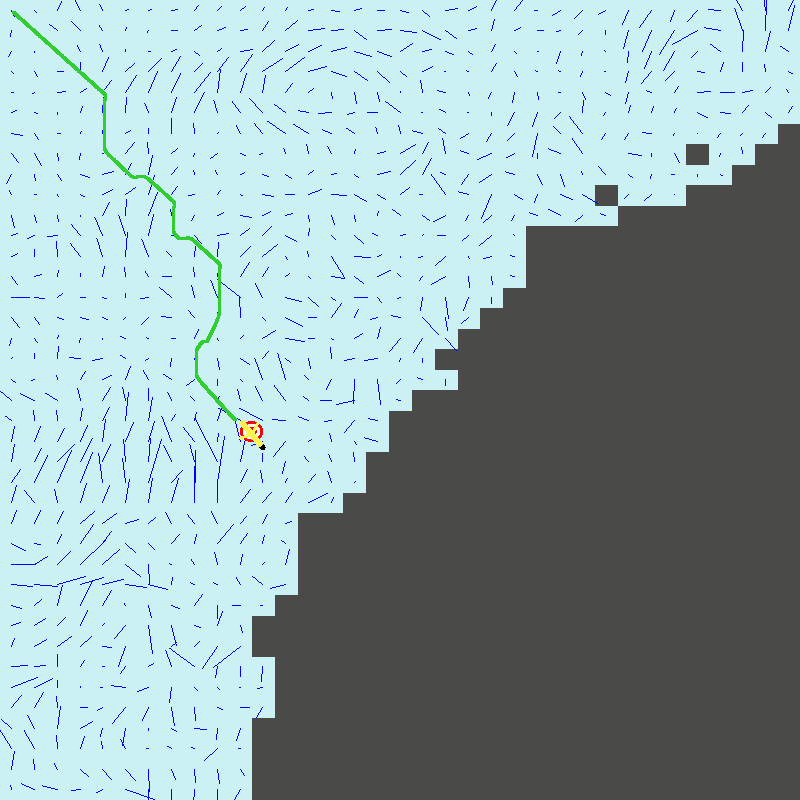}}
    \qquad
    \subfigure[]{\label{fig:iterative_real}\includegraphics[width=0.2\textwidth]{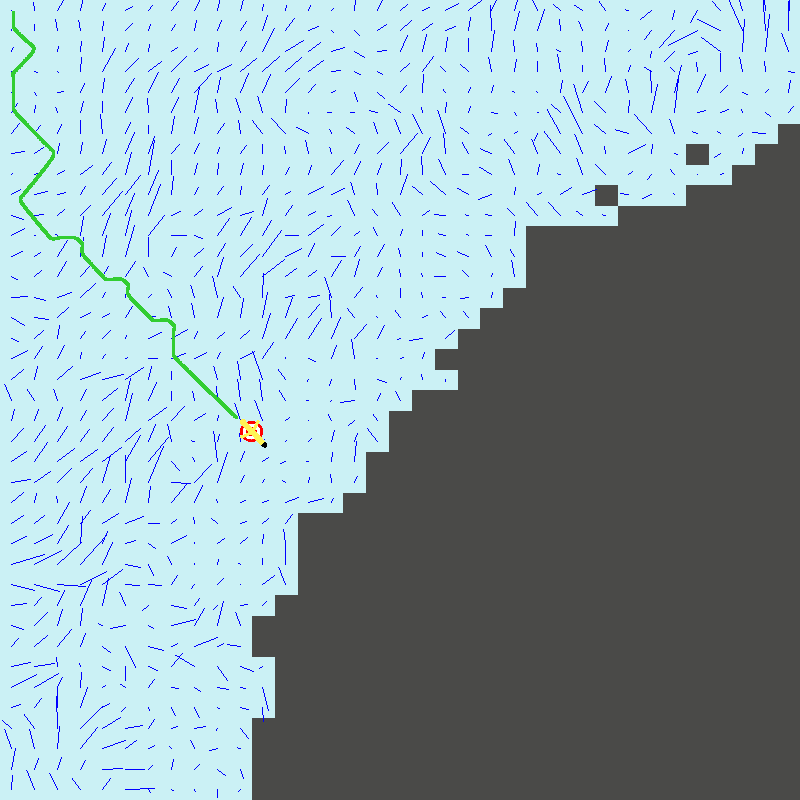}}
    \qquad
    \subfigure[]{\label{fig:tvmdp_vi_real}\includegraphics[width=0.2\textwidth]{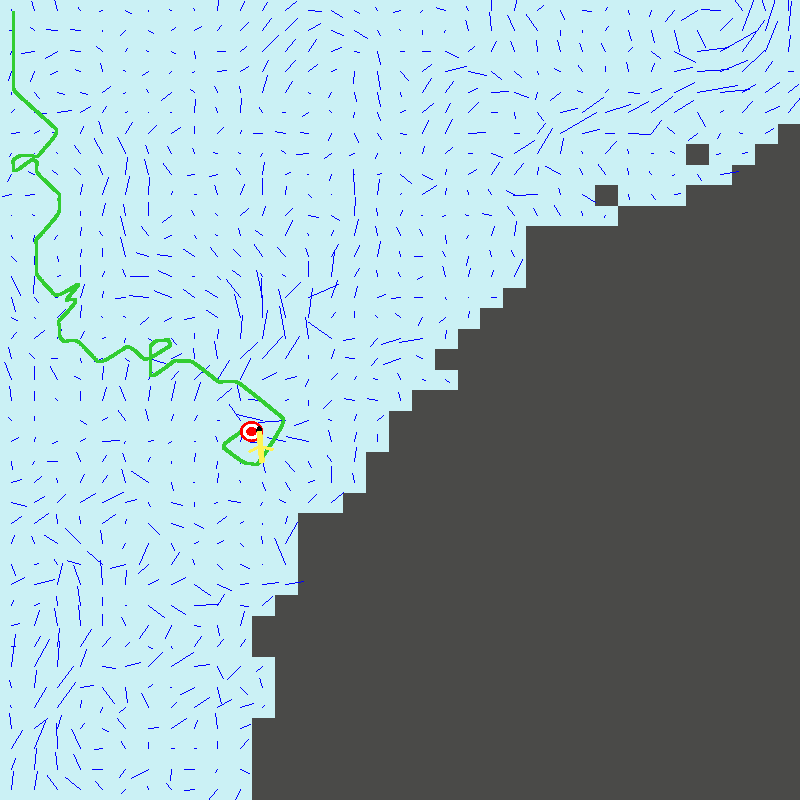}}
    \vspace{-15pt}
    \caption{\small Trajectory results using the entire ROMS data. (a) Trajectory from the full spatiotemporal space; (b) Trajectory from Alg.~\ref{alg:vi_reduced}; (c) Trajectory from Alg~ \ref{alg:vi_mfpt}.}
    \label{fig:real_trajectories}
\end{figure*}

We start with a simplified setting without considering the robot kinematics model: we set local transition time $h(s, t, s') = 1$ for all $s, s' \in \mathbb{S}$ and $t \in \mathbb{T}$. 
Such a simplified model allows us to obtain the optimal policy as a ground truth. 
Specifically, 
we discretize the time into 50 slots with a specified horizon of 50 transitions, and 
the spatial state $\mathbb{S}$ is represented by a $13 \times 13$ grid plane. 
Therefore, the full size of our spatiotemporal state space is $13\times13 \times 50= 8450$.
We use a Gaussian distribution to model the robot's state transition function where the variance is set to be $0.6$. We set $A=4, \omega=1$ for self-spinning model.
The parameters of the dynamic vortex model are set to $r=3, \omega=1$.
For ROMS data, the same size of spatiotemporal state space is also chosen (cropped) from the raw dataset. 
Fig.~\ref{fig:toy_exp} shows the average number of state transitions of the four algorithms, 
where the number of state transitions implies the 
number of hops used to reach the goal state (the smaller the number, the better the algorithm). 
The VI in the full space $\mathbb{S}\times\mathbb{T}$ (with dotted yellow texture) is the exhaustive search in the entire space, thus it provides the optimal solution.
It is obvious that Alg.~\ref{alg:vi_reduced} (in red stripe) has a performance that is the closest to the optimal. 
The non-iterative method gives a worse performance, indicating that the 
iterative procedure helps improve results. 

Fig.~\ref{fig:num_states} illustrates the effectiveness of the state reduction by Alg.~\ref{alg:vi_reduced}. 
It shows the size of the reduced state space at each iteration as well as the total number of states visited by the algorithm.
One can observe that the number of states visited at each iteration is around one third of the full space.

Fig.~\ref{fig:toy_traj} shows the trajectories of the four methods in two types of disturbance vector fields, 
which  reveal that our algorithm achieves the closest performance to the optimal solution.

Statistics on computational costs per iteration are shown in Fig.~\ref{fig:time_complexity}. 
Since the reconstruction causes the reachable spatiotemporal space to change at every iteration, we average the computational time over all iterations.
It is obvious that our method (Alg.~\ref{alg:vi_reduced}) requires much less computational time than the VI in full space.
The gap between our method and non-iterative algorithms is mainly due to the overhead in computing the first and second moments of PPTs.  
Since the linear systems corresponding to first and second moments of PPTs are separate, 
multi-threading techniques can be utilized to further decrease the computational cost without affecting the final result.
Fig.~\ref{fig:multi_thread} shows the improvement using multi-thread computation.

{
    \begin{figure}[t]
        \vspace{-15pt}
        \begin{subfigure}[]
        {\label{fig:traj_real}\includegraphics[scale=0.3]{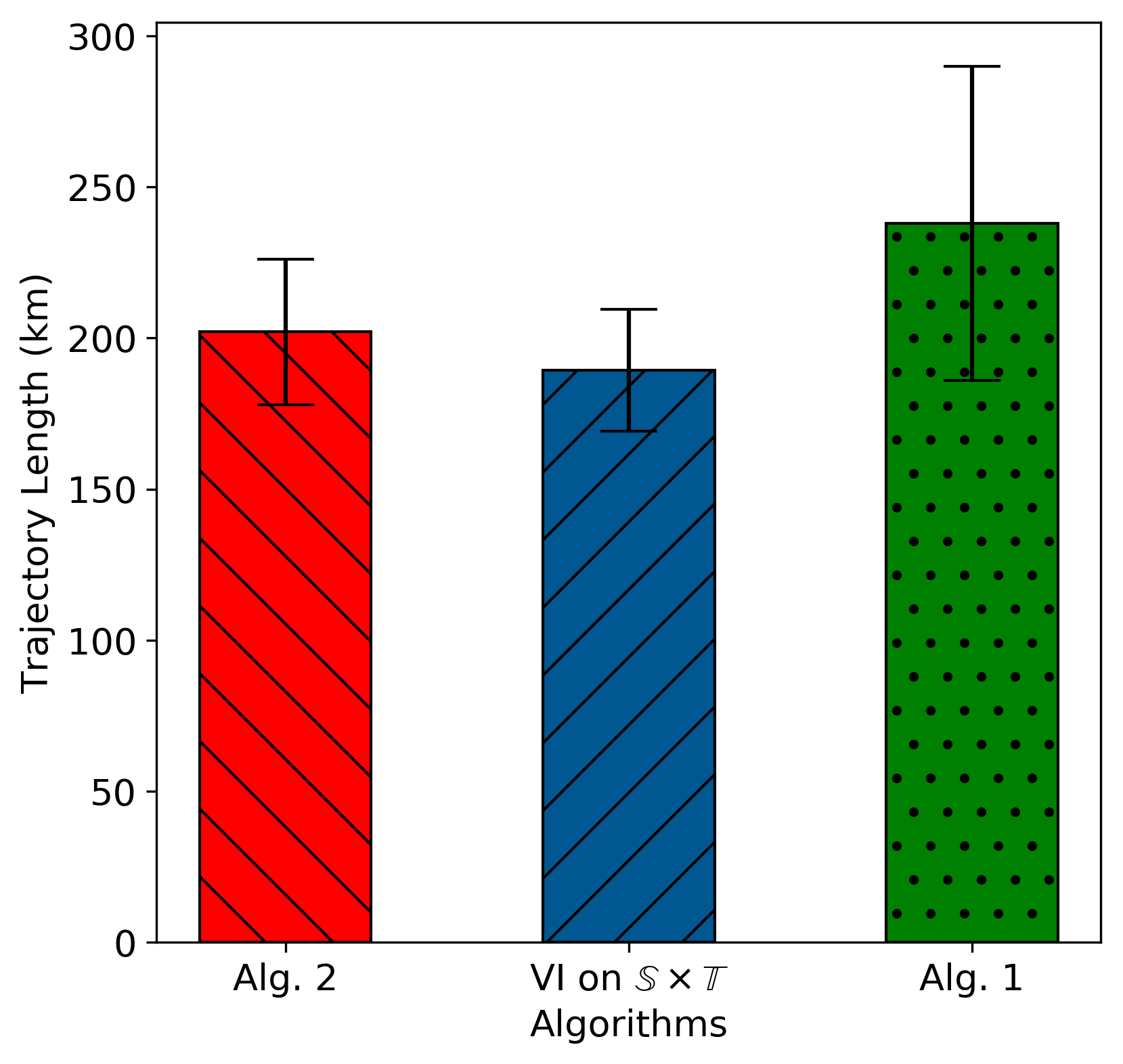}}
        \end{subfigure}%
        \begin{subfigure}[]
        {\label{fig:time_real}\includegraphics[scale=0.3]{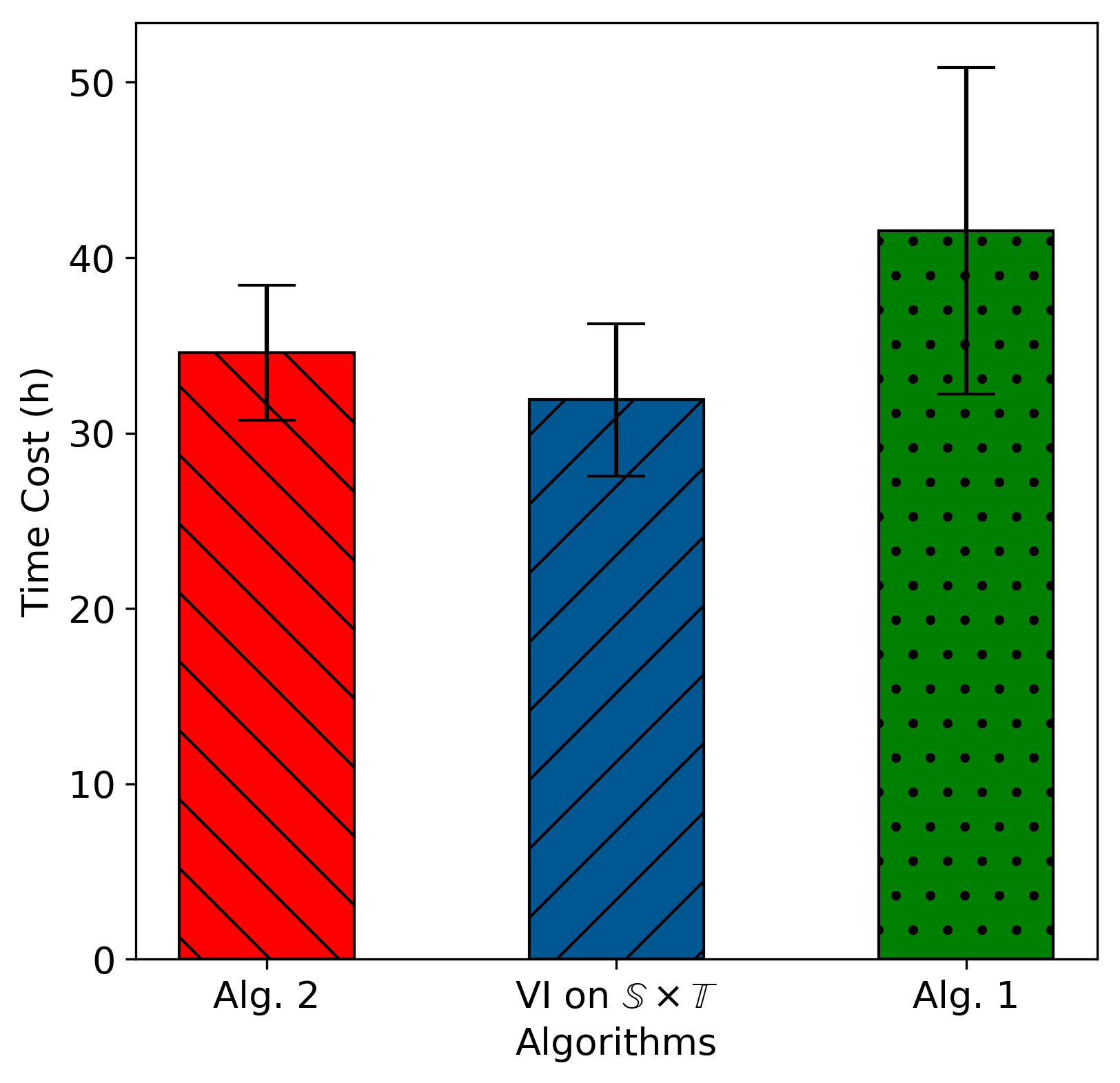} \vspace{-10pt}}
        \end{subfigure} \vspace{-15pt}
        \caption{\small (a) Trajectory lengths of the three algorithms; (b) Time costs. The statistics are averaged over 100 trials.  \vspace{-15pt} }
        \label{fig:real_performance}
    \end{figure}
}

\subsection{Realistic and High-Fidelity Scenarios}\label{sec:realistic}
In this experiment, 
we consider the task of navigating an autonomous underwater vehicle to a goal state with a kinematics model.
Specifically, a PID controller is used to follow the high-level action $a = \pi(s, t)$ generated by the policy given the current state-time pair $(s, t)$. 
In this case, the local transition time $h(s, t, s')$ is no longer a constant because it is influenced by the vehicle motion dynamics and the time-varying ocean disturbances.
We evaluate the algorithms using the entire region of the ROMS data, 
which is discretized into a $35 \times 39$ grid map with a spatial resolution of $\mbox{6 km} \times \mbox{6 km}$.
The vehicle's maximum velocity is set to $\mbox{6 km/h}$.


In order to solve the first and second moments of PPTs, we need to compute the local transition times first.
However, the exact form of $h(s, t, s')$ is hard to obtain due to the continuous vehicle motions, ocean currents, and the state space.  
To address this problem, we resort to a method that utilizes Monte Carlo trials to estimate $h(s, t, s')$.

Specifically, we want to approximate the traveling time from state $s$ at time $t$ to its next state $s'$ given the vehicle motion and the current ocean disturbance estimated via GPR.  
We first discretize the orientation of vehicle's motion as well as the ocean disturbance direction into eight directions. Then, we sample the next state $s'$ randomly from the transition function
for each pair of discrete motion and disturbance directions. 
The local transition time is calculated as $h_a(s, t, s')=\frac{v_{net}}{d}$, 
where $v_{net}$ is the net 
velocity of the vehicle
after taking account of the ocean disturbance at time $t$
and $d$ is the distance between $s$ and $s'$. 
To obtain an estimate of local transition times for all states, 
the above method iterates over the entire 
spatiotemporal state space with 100 trials for each state. 
Note that this estimation can be computed offline without adding any computational cost to the proposed algorithms.


\begin {table}[t]
\centering
\vspace{-10pt}
\caption {\small Computational cost for one iteration using the whole original ROMS data over three days.}
\label{tb:real_computation}
    \resizebox{1\linewidth}{!}
    {    \vspace{-10pt}
         \begin{tabular}{|c|c|c|c|c|}
         \hline
         Alg.~\ref{alg:vi_reduced} & Alg. \ref{alg:vi_mfpt} & \begin{tabular}{@{}c@{}}VI in \\ $\mathbb{S}\times\mathbb{T}$\end{tabular} & \begin{tabular}{@{}c@{}}Multi-Thread\\ Alg.~\ref{alg:vi_reduced}\end{tabular} & \begin{tabular}{@{}c@{}}Multi-Thread\\ Alg. \ref{alg:vi_mfpt}\end{tabular} \\
         \hline
            3min & 1min & 15min & 2.4min & 0.3min\\
        \hline         
        \end{tabular} \vspace{-15pt}
    }
\end{table}

We use the above estimation of local transition time 
to compute the first and second moments of PPTs
in Alg.~\ref{alg:vi_mfpt} and  Alg.~\ref{alg:vi_reduced}.
Because of the high computational cost of VI in $\mathbb{S}\times\mathbb{T}$, 
we choose to discretize the time dimension with $30$ slots with a time horizon of $50$ 
hours.
To keep the comparison fair, we choose the same discretization for Alg.~\ref{alg:vi_reduced}.
We exclude the non-iterative method in this experiment as it does not produce comparable performance.

Fig.~\ref{fig:real_performance} shows the statistics of the three algorithms in terms of trajectory length and time cost.
The results reveal similar performance for scenarios described in the previous section, except that the performance gap between VI in $\mathbb{S}\times\mathbb{T}$ and Alg.~\ref{alg:vi_reduced} is slightly smaller. 
This is because the discrete time intervals in Alg.~\ref{alg:vi_reduced} are adjusted according to the local transition time, as mentioned in the end of Section~\ref{sec: ST-Space}.
The corresponding trajectories are shown in Fig. \ref{fig:real_trajectories}. 
The computational time is presented in Table \ref{tb:real_computation}. 
We can see that, in this test scenario our algorithm requires only one-fifth of the time used by the VI over the full spatiotemporal space.


\vspace{-3.5pt}
\section{Conclusions}
\vspace{-1pt}
We present a passage percolation time-based method to time-varying Markov Decision Processes that can be applied to autonomous systems'  motion (action) planning. 
Our method iteratively solves the TVMDP reconstructed from a reachable space that was reduced from the original state  space based on the first and second moments of passage percolation time. 
Our extensive simulations using ocean data show  
that our approach produces results close to the optimal solution but requires much smaller computational time.
We will incorporate the methods of statistical learning for data into our proposed framework in the future.







\newpage

\bibliographystyle{plainnat}
\bibliography{ref.bib}

\end{document}